**POLYTECHNIQUE MONTRÉAL**

affiliée à l'Université de Montréal

**Multi-Agent Language Models: Advancing Cooperation, Coordination, and Adaptation**

**ARJUN VAITHILINGAM SUDHAKAR**

Département de génie informatique et génie logiciel

Thèse présentée en vue de l'obtention du diplôme de *Philosophiæ Doctor*
Génie informatique

April 2024

**POLYTECHNIQUE MONTRÉAL**

affiliée à l'Université de Montréal

Cette thèse intitulée :

**Multi-Agent Language Models: Advancing Cooperation, Coordination, and Adaptation**

présentée par **Arjun VAITHILINGAM SUDHAKAR**
en vue de l'obtention du diplôme de *Philosophiæ Doctor*
a été dûment acceptée par le jury d'examen constitué de :

**Chris PAL**, président
**Sarath CHANDAR**, membre et directeur de recherche
**Siva REDDY**, membre externe

# TABLE OF CONTENTS

# LIST OF TABLES

# LIST OF FIGURES

# LIST OF APPENDICES

# CHAPTER 1    INTRODUCTION

Modern Large Language Models (LLMs) [2–4] demonstrate remarkable zero-shot or few-shot generalization capabilities and complex natural language tasks [5]. Additionally, the ability to understand and generate human-like text has made them the perfect choice as virtual assistants for a wide variety of tasks, ranging from language translation [6] to text summarization [7]. It is particularly striking that these models seem to understand the meaning behind the textual interactions they encounter while only having been trained on a large corpus of text documents, with no direct evidence of the intentions of their authors. This gives reason to believe that LLMs are powerful enough to infer the meaning behind spoken words, a question that researchers have tackled for many decades [8–11]

Understanding the meaning behind other people's actions and being able to infer what they will do in the future is key towards collaborative success [12]. One could even argue that the reason humans were able to thrive as a society is due to their ability to cooperate. These interactions among multiple agents are essential components of various aspects of our lives, ranging from everyday activities like commuting to work to functioning fundamental institutions like governments and economic markets. Effective collaboration and coordination with humans or other autonomous systems become imperative as human-artificial interactions continue to increase.

Although LLMs seem to understand the meaning behind spoken words, it remains to be seen if they can go beyond understanding and appropriately model and act upon people's intentions. Do they possess *theory of mind* [13,14], i.e., can they reason about the world from another agent's perspective? Answering this question is of crucial importance for a successful collaborative future, as LLMs will become increasingly intertwined with our lives. We will analyze the theory of mind of LLMs as a cooperative multi-agent reinforcement learning (MARL) problem, where artificial agents learn to collaborate together through repeated interactions in the same principled manner as humans would.

Through this proposal, we aim to significantly advance the adaptation capabilities of artificial agents towards others, both when paired with other artificial agents or humans. The ability to cooperate with other agents and humans is a desirable characteristic of an interactive agent that learns in the presence of other agents and is a fundamental capability to thrive as a society. We push towards this goal through LLM-based agents, with whom we already regularly interact with [2–4, 15], and which are capable of producing natural language. As LLMs are everywhere and will continue to spread, and we increasingly interact with them, the outcomes of this proposal have the potential to significantly impact society, creating truly hybrid systems.

## 1.1   Research Goals

In this proposal, we aim to work towards the following three goals:

Goal 1   **Benchmarking theory of mind capabilities of current LLMs**. LLMs have been trained on a vast corpus of text and have acquired such diverse general knowledge that they can extrapolate toward unseen tasks. We aim to benchmark on large models and analyze how well this general knowledge extends to reasoning about other's behaviors. This involves assessing how well LLMs can understand and participate in cooperative play with other LLMs by interpreting their intentions, utilizing the Theory of Mind, and predicting their moves.

Goal 2   **Knowledge transfer among games with different number of agents.** MARL-Hanabi's research has concentrated on the two-player version, and scaling to 3, 4, or 5 players requires billions of interactions and limited performance [12]. Such large-scale RL systems require restarting learning them from scratch when moved from 2 players to a 3, 4, or 5-player setting, which may require weeks if not months of computation, and there may be billions of data points to re-process [16]. Hence, this makes the tabula rasa approach impractical and a significant bottleneck in training generalist agents. The RL policies of the 2-player game cannot be transferred to 3 players due to different dynamics/strategies and changing states & action spaces, which limits the general purpose agents. One significant strength of LLMs is its ability to transfer its knowledge to diverse downstream tasks. We aim to build architectures with LLM as a backbone to leverage existing knowledge and computations. Using these learned network weights, to accelerate training and increase performance, transfer knowledge between Hanabi games with varying numbers of players.

Goal 3   **Leverage the LLM's few-shot reasoning capabilities to adapt to novel partners at test time quickly.** Agents using different learning algorithms will converge to various strategies and make different assumptions about what their partners will do. When paired together, these agents will underperform [17]. Even more striking is that the same learning algorithms trained in a 2-player game will fail to play together in a 3-player game. This is a fundamental limitation exhibited by many of the current MARL algorithms. In a real-world setting, completely different types of agents should be able to be paired together and succeed in collaborative tasks. We will develop few-shot coordination techniques, taking into account the theory of mind, such that our agents can cope with unseen partners or adapt to changes in strategy. Few-shot coordination is a more practical approach to the zero-shot coordination paradigm recently explored by the MARL community [18], as we cannot predict every type of agent we will encounter in our lifetime.

In pursuing goals Goal 1, Goal 2, Goal 3, we investigate the following crucial problem statements essential to this proposal.

In Chapter 3, we aim to develop a generative model that can rapidly adapt with minimal reliance on human gameplay datasets while matching or surpassing the existing baselines. These capabilities have practical applications in both single-agent and multi-agent setups, enhancing the efficiency, transfer and adaptability of the agents.

In Chapter 4, our objective is to establish a challenging benchmark for Multi-Agent language models. The current benchmark for the Multi-Agent language model falls short in scale or complexity to test [19], or it is an open world [20], which makes it hard to evaluate generative agents. However, in the domain of MARL, the problem of coordination and cooperation is extensively explored as a central focus [12]. Hence, combining MARL methodologies with LLM abilities seems like a promising avenue toward achieving long-term collaborative success and developing artificial agents that develop a theory of mind.

# CHAPTER 2    BACKGROUND

In this chapter, we provide the relevant background to understand the fundamentals related to language models and reinforcement learning in this section that are essential to understanding the rest of the thesis proposal.

## 2.1    Language Model

Language models learn from vast corpora of text data in an unsupervised manner to capture the distribution of words and phrases, thereby uncovering underlying statistical patterns and distributions [21]. These models can predict the likelihood of a sequence of words or create new sequences that are both syntactically and semantically coherent, given enough parameters of the model and the dataset used in the training process [22].

The introduction of Transformer models [23], has enabled parallelized training, large-scale advancements in both model and dataset sizes in language models. This breakthrough was primarily attributed to the use of attention mechanisms [24]. These mechanisms allocate weights to input tokens according to their significance, thereby enabling the model to prioritize more relevant tokens. Attention in transformers [23] computes query, key, and value mappings for input sequences by projections.

$$\text{Attention}(Q, K, V) = \text{softmax}\left(\frac{QK^T}{\sqrt{d_k}}\right) V \tag{2.1}$$

This equation represents the scaled dot-product attention, where $Q$, $K$, and $V$ stand for the query, key, and value matrices respectively. $d_k$ represents the dimensionality of the keys. Scaling by $\sqrt{d_k}$ prevents the dot products from growing too large in magnitude.

The different attention strategies used in these transformers are,

**Self-Attention:** Calculates attention using queries, keys, and values from the same component (either the encoder or decoder).

**Cross Attention:** Employed in the encoder-decoder architecture, where the outputs of the encoder serve as queries, and the key-value pairs originate from the decoder.

Encoder-decoder-based transformers [23] process inputs through the encoder and pass the intermediate representation to the decoder to generate the output. In this architecture, the encoder processes the complete sequence utilizing self-attention, whereas the decoder processes the sequence sequentially, implementing cross-attention.

### 2.1.1 Encoder Based Model

By contextualizing input data, encoder-based models [25] turns raw text into a more abstract representation of its meaning. This process involves analyzing the text to understand its syntax and semantics, thereby encoding the essential information into a fixed-length vector regardless of the input's length. These models are well suited for tasks like text classification [26], sentiment analysis [27] etc.

**BERT** [25] is a powerful transformer-based model designed for natural language processing tasks. It is pre-trained on large corpora and captures bidirectional context information, allowing it to understand context-dependent meanings of words in a sentence.

**DistilBERT** [28] is a distilled version of BERT, designed to retain most of its performance while being computationally more efficient. It achieves this by reducing the number of parameters and layers.

The model's finetuning process begins with a set of training instances, denoted as $(S, A)$ drawn from the dataset $\mathbb{D}$ where $S \in \{s_0, s_1, .., s_n\}$ and $A \in \{a_0, a_1, .., a_n\}$. Within this set, $s$ and $a$ represent state and its corresponding labelled output action, respectively, and $n$ represents the number of examples in the dataset.

The finetuning objective of BERT, DistilBERT is as follows,

$$L_{CCE} = -\frac{1}{N} \sum_{i=1}^{N} \sum_{j=1}^{C} a_{ij} \log(\hat{a}_{ij}) \tag{2.2}$$

Where $N$ is the batch size. $C$ is the number of classes. $a_{ij}$ is the true probability of class j for the i-th example in the batch and $\hat{a}_{ij}$ is the predicted probability of class j for the i-th example in the batch.

### 2.1.2 Decoder Based Model

In contrast, decoder-based models specialize in generating text [29]. Starting from an encoded input or a given context, these models focus on producing a sequence of tokens (words or characters) one at a time [30–32]. Decoder models excel in tasks like text generation, including creative writing, code generation, and more. They predict the next token in a sequence based on the previous tokens, effectively *decoding* the context or input into a coherent output.

**GPT-2 Generative** [30] language model known for its capacity to generate coherent and relevant text which acts as a generator.

The finetuning objective of GPT-2 Generative is to minimize the cross-entropy loss, denoted as $\mathcal{L}$. The cross-entropy loss is mathematically defined as follows:

$$\mathcal{L}_{LLM} = -\mathbb{E}_{(S,A) \sim D} \log p(A|S) \tag{2.3}$$

Where $p(S|A)$ represents the conditional probability of predicting an action $A$, given the state $S$. The goal is to optimize these parameters, by minimizing the cross-entropy loss.

**GPT-2 Classifier** [30], we add a linear layer on top of pre-trained layers of the decoders model with its output dimension equaling our number of labels.

## 2.2   Reinforcement Learning

Reinforcement Learning (RL) is a machine learning paradigm focusing on problem-solving through decision-making [1]. In this paradigm, an agent interacts with its surroundings, earning rewards or penalties depending on its actions. RL has been successfully applied in various domains, including gaming [33], robotics [34], and the development of self-driving cars [35], demonstrating its versatility and potential for addressing complex decision-making problems.

The core problem in reinforcement learning (RL) involves an agent's learning behaviors to achieve goals through interactions with an environment. The agent receives information about the state from the environment, performs actions in the environment, and receives scalar rewards as a feedback signal. This process is depicted in Figure 2.1

The notation used in RL is as follows,

- Agent: The learner or decision-maker.

- Environment: The system with which the agent interacts.

- State ($s$): A representation of the status of the environment.

- Action ($a$): An operation performed by the agent on the environment.

- Reward ($r$): A scalar feedback signal indicating the immediate benefit of taking an action in a state.

- Policy ($\pi$): A strategy for choosing actions based on states.

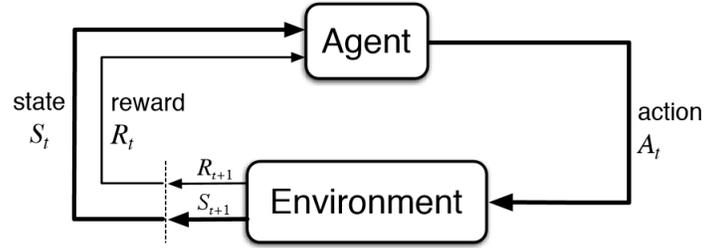

Figure 2.1 The interaction between an agent and the environment in RL [1].

### 2.2.1 Markov Decision Processes (MDP)

The framework of Markov Decision Processes (MDP) [1] provides a mathematical formalization for sequential decision-making problems that RL aims to solve. An MDP is defined by the tuple $(S, A, P, R, \gamma)$, where:

- $S$: A set of states.

- $A$: A set of actions.

- $P$: State transition probability matrix, $P(s_{t+1}|s_t, a_t)$.

- $R$: Reward function, $R(s_t, a_t)$.

- $\gamma$: Discount factor, $0 \leq \gamma \leq 1$.

The overall objective is to find a optimal policy $\pi$ that maximizes the expected cumulative reward.

- Value Function: $V^\pi(s) = \mathbb{E}[\sum_{t=0}^{\infty} \gamma^t R_{t+1}|S_0 = s, \pi]$.

- Action-Value Function: $Q^\pi(s, a) = \mathbb{E}[\sum_{t=0}^{\infty} \gamma^t R_{t+1}|S_0 = s, A_0 = a, \pi]$.

### 2.2.2 Partially Observable Markov Decision Processes (POMDPs)

The traditional MDPs can be extended to scenarios when the agent is unable to completely view the state of the environment by Partially Observable Markov Decision Processes (POMDPs) [36]. In POMDPs, the agent does not have direct access to the state $s$ but instead receives observations $o$ that provide partial information about the state. Therefore, depending on the history of actions done and observations received, the agent's decision-making process must rely on a *belief state*, a probability distribution over all potential states. POMDPs are defined by the tuple $(S, A, P, R, \Omega, O, \gamma)$, where:

- $\Omega$: A set of observations that the agent can receive.

- $O$: The observation function, $O(o|s', a)$, which defines the probability of receiving observation $o$ after taking action $a$ and ending up in state $s'$.

### 2.2.3 Deep Q-Network (DQN)

Using tabular methods works when the size of $S \times A$ is relatively small, but when the size of that set becomes large or even infinite, its necessary to use function approximation to represent the policy or value estimates. [37] introduced Deep Q-Networks (DQNs) to play Atari games, showing that deep learning could be used in RL. DQN is a version of Q-learning that used deep neural networks to represent the Q-function. These networks are then trained using gradient based optimization algorithms with a loss based on the Bellman equations.By approximating the Q-value function with a deep neural network, DQNs can efficiently handle complex environments that are challenging for traditional Q-learning.

Q-learning seeks to learn the action-value function $Q(s, a)$, which estimates the total expected rewards for taking an action $a$ in a state $s$ and following the optimal policy thereafter. The Q-learning update rule is given by:

$$Q(s_t, a_t) \leftarrow Q(s_t, a_t) + \alpha \left[ r_{t+1} + \gamma \max_a Q(s_{t+1}, a) - Q(s_t, a_t) \right] \tag{2.4}$$

DQN introduces the use of experience replay and fixed Q-targets, which address the issues of correlated data and moving targets in the learning process. Thus leading to more stable & effective training and handling the deadly triads (Off-Policy, Boostrapping, Approximation).

### 2.2.4 Deep Relevence Recurrent Network (DRRN)

A prominent deep reinforcement learning technique in text-based games is the Deep Reinforcement Relevance Network (DRRN), a variant of the Deep Q-Network (DQN). DRRN is mainly used because it is robust in handling large and dynamic action spaces characteristic of text environments [38]. The observation ($o$) and actions ($a$) are first encoded using separate recurrent neural network encoders (such as a GRU [39]) $f_o$ and $f_a$ respectively. A decoder $g$ then combines the representations to obtain the Q-value using a network parameterized by $\Phi$:

$$Q^{\Phi}(o, a) = g(f_o(o), f_a(a)). \tag{2.5}$$

The DRRN learns to estimate the Q-value through iteratively updating $\Phi$ with experience sampled from a prioritized experience replay buffer with the temporal difference (TD) loss [1]:

$$\mathcal{L}_{TD}(\Phi) = \left( r + \gamma \max_{a' \in A} Q^{\Phi}(o', a') - Q^{\Phi}(o, a) \right)^2,$$ (2.6)

where $r$ and $o'$ are the reward and the observation received after taking action $a$ upon observing $o$, and $\gamma$ represents the discount factor.

### 2.2.5 Multi-Agent Reinforcement Learning (MARL)

Multi-agent reinforcement learning (MARL) is a branch of machine learning that extends the traditional reinforcement learning (RL) paradigm to scenarios involving multiple agents interacting within a shared environment. In MARL, each agent aims to learn an optimal policy that maximizes its cumulative reward over time, while taking into account the actions and policies of other agents. This complexity introduces unique challenges, such as the non-stationarity of the environment from the perspective of any single agent, as the strategies of other agents evolve concurrently. The dynamics of multi-agent interactions can be formalized through game theory, where the Nash equilibrium often serves as a goal for learning stable strategies.

The collective actions of all agents at a given time form the action profile $\mathbf{a} = (a_1, \ldots, a_N)$, with $N$ being the number of agents. The transition function $\mathcal{T} : \mathcal{S} \times \mathcal{A}_1 \times \ldots \times \mathcal{A}_N \rightarrow \mathcal{S}$ determines the next state of the environment based on the current state and the action profile. Each agent has its own reward function $R_i : \mathcal{S} \times \mathcal{A}_i \rightarrow \mathbb{R}$, emphasizing that agents may have different objectives. The goal of each agent is to maximize its expected cumulative reward, often represented as $E\left[ \sum_{t=0}^{\infty} \gamma^t R_i(s_t, a_{i,t}) \right]$,

The study of MARL not only seeks to develop algorithms that allow agents to learn efficiently in the presence of other learning agents but also to understand the fundamental properties of multi-agent interaction, cooperation, and competition. This research area is pertinent to a broad range of applications, including autonomous vehicles, distributed control systems, and strategic games, making it a critical domain within artificial intelligence.

## 2.3 Text Games

Text-based games are interactive simulations in which the game state is in text, and the player has to progress by providing text-based commands [40]. The player's objective is to maximize the game score by understanding the game's objective and solving it through sequential decision-making. In text-based games, language has been used as an instruction for a reinforcement learning agent in a partially observable Markov Decision Process (POMDP). TextWorld has proven to be a useful testbed for developing agents that operate in language. Working on this problem is interesting and challenging, as interactions in these games (input observations, action commands) are done through text. The agent will require solid language understanding for successful gameplay.

Compared to TextWorld, Jericho [33] have significantly more linguistic variety and larger action space. These learning environment supports a collection of 32 human-written interactive fiction games. These games are designed to be difficult for human players, serving as a more realistic training ground for intelligent agents. The most commonly used reinforcement learning agent is the Deep Reinforcement Relevance Network (DRRN) [41] for baselines to handle dynamic action space and robust to changes in action space.

The learning agent interacts with the game environment at each time step $t$ by carrying out a textual action $a_t$ from a set of feasible actions $\mathcal{A}_t$, depending on the textual observation $o_t$ that is currently being made. The reward is obtained from the environment for every action, which guides the learning agent. The overall goal of the agent is to maximize the overall return of the game $r_t = \mathcal{R}_t(o_t, a_t)$. The episode ends when the agent completes the maximum number of steps at any given time $t$, dies, or accomplishes its purpose.

**CHAPTER 3    Language Model in the Loop: Towards Adaptation**

This chapter is a slightly modified version of our paper [42].

**My contributions as First Author:** Problem statement identification, literature survey, benchmark establishment, implementation and running experiments.

**Status:** The paper is currently under at arXiv 2023 under the title "Language Model-In-The-Loop: Data Optimal Approach to Learn-To-Recommend Actions in Text Games".

## 3.1    Introduction

Large Language models [25,43] (LLMs) trained on large corpora of unstructured text corpora are the state-of-the-art models in several Natural Language Understanding (NLU) benchmarks. [44] argue in their position paper that the models trained largely from static benchmarks rely to the *form* rather than understanding the meaning. While it is imperative to understand the learning dynamics of LLMs [45,46], introducing novel language understanding challenges pushes the frontiers for LLMs' applications.

There has been a recent interest in interactive training of large language models in situated learning environments. [47,48] point out the necessity for LMs to have enhanced language understanding and meaning through interacting with the physical world. Also, [49] argues that LMs fall short in their communicative usage, requiring reasoning over intents despite their success in static datasets.

Training decision making agents over textual information for playing text-based games [33,40] has been a recent usecase for LLM. While decision making has been the front of text-game playing, such games introduce novel challenges for language understanding, and domain adaptation for LLMs. [50] used GPT-2 [43] to generate candidate actions for the decision making DRRN module [41] in Jericho benchmark of text based games. Such a set up allows for qualitatively understanding the LLMs' abilities to *understand*, *reason*, and *adapt* to novel situations. In a typical text-based game, as in Figure 3.1, an agent receives a textual observation about its environment that it has to understand and reason over the possible actions to pick one and proceed. While learning from scratch is time-consuming, [50] make use of linguistic priors in LLMs to prune the combinatorially large action space. The authors adapt GPT-2 for the task with a corpus of human game play on similar games— ClubFloyd. After the adaptation phase, the model remains frozen throughout the learning that happens within the game.

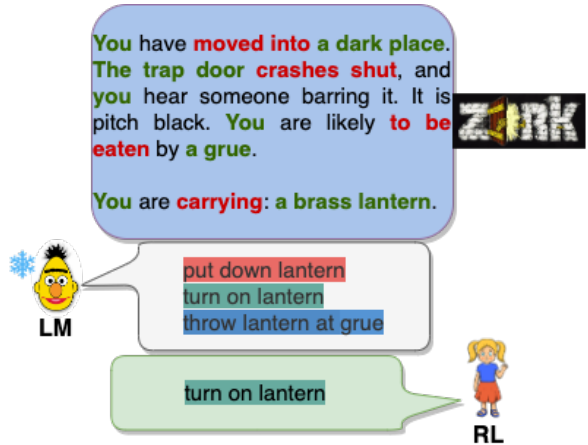

Figure 3.1 Sample gameplay from zork1 game in Jericho using LM for action recommendation: LM recommends action candidates based on the observation from `env`. The RL agent selects an action from the candidates.

Further, [50] also note that the performance on the text-based games in Jericho benchmark was sensitive to the size of the annotated human gameplay corpus; such reliance adds to the cost. On the one hand in-game transitions remain unutilized for training the LLM, and on the other there is a need to mitigate the reliance on human annotated transitions to scale applications of LLMs. Although one can make use of the transitions to train the model, the solution requires a comprehensive analysis on what such a LM-in-the-Loop training entails.

Toward that, we explore LM-in-the-Loop by building over the setup in [50] by training GPT-2 using in-game generated transitions. Further, we analyze such a set up along the metrics of: (1) Improvement in performance, (2) Acceleration in convergence, (3) Reliance on human annotated transitions, (4) comparing transitions selection for LM training,

The main findings of the approach are summarized as follows:

- LM-in-the-Loop reduces emphasis on human-annotated transitions

- Improves the existing state-of-the-art approach and enables accelerated convergence.

- State feature-based transitions selection provided more significant gains than other alternates.

## 3.2 Related Work

**Text Games:** Jericho [33] is a popular learning environment that supports 32 human-written interactive fiction games. These games are designed to be difficult for human players, serving as a more realistic training ground to evaluate language understanding agents. Compared with frameworks like TextWorld [40], these games have significantly more linguistic variety and larger action space. Jericho environment provides a smaller list of candidate actions that can be used to train reinforcement learning (RL) agents. Approaches like DRRN [41], TDQN [33], and KGA2C [51] have used *handicap* to operate on small action space and learn only through in-game rewards. Towards using large LMs, environment provided actions are replaced with LM generated actions like with GPT-2 [50], or BERT [52].

**Transformers in RL:** Transformer architectures are now being increasingly used in reinforcement learning (RL); [53, 54] use smaller transformer architectures on Atari games that earlier used convolutional networks as policy networks in offline setting. Further adaptations to make the architectures lightweight to enable online training was proposed in [55–60]. [50] explore using the semantic prior in GPT-2 for candidate action recommendation in text games. Further, [61,62] train LMs to remember optimal trajectories to swiftly move to novel game regions.

**Data Efficiency:** LLMs [63] are pretrained with tremendous amount of unstructured text data from *the web* using a generic language modeling objective. Adapting the models to a downstream tasks [26,27,64,65], however, has been shown to greatly affected by the quality of supervision and the size of the dataset. As reliance on annotated data makes their application hard to scale, techniques like data augmentation [66], using distilled models [67], learning from toyish data [68] has been explored has alternatives. However, the approach of making LLMs interactive to be trained in a situated learning environment to reduce the need for annotations is only recently getting popular.

## 3.3 Methodology:

### 3.3.1 LLM for Action Recommendation

Consider a dataset $\mathcal{D}$ of $N$ transitions of human gameplay across different games organized in context-action pairs as $((o_{j-1}, a_{j-1}, o_j), a_j)$. For example: a sample could be like, "`[CLS]`... to the north is a restaurant where the mayor ate often. to the east is the mayor's home. `[SEP]` northeast `[SEP]` ... you are carrying nothing. you are still on the streets. ... `[SEP]` northeast". `[SEP]` and `[CLS]` are special tokens specific to LM-training. [50] uses ClubFloyd to adapt a pretrained GPT-2 model with causal language modeling task. The motivation is to enable the linguistic prior of GPT-2 to adapt to the games and provide better action recommendations to the DRRN.

### 3.3.2 LM-in-the-Loop to recommend Actions

The game playing agent follows trajectories that are rewarded according to the rules of the game in the Jericho environment. The environment has two scenarios—with and without *handicap*—which correspond to whether the actions can be generated from within the possible actions suggested by the environment or without any limitations by the environment respectively. The *with handicap* set up evaluates the agent exclusively on planning with the actions provided, while the *without handicap* requires the agent in addition to understanding the observation also generate acceptable candidates. In [50], the LLM is kept constant throughout the gameplay and that assumption could be only validated if Jericho games share significant similarity with the transitions in ClubFloyd.

Toward that, we explore the feasibility, prospects, and challenges that entail training LM-in-the-loop post finetuning with human gameplays in ClubFloyd adaptation as in Table 3.1. We use a similar set up for action recommendation as in [50], where a pretrained GPT-2 LM is adapted with Clubfloyd dataset to recommend actions to DRRN agent. In addition to training the DRRN agent with TD-learning (Equation 2.6), we collect the transitions $(o_t, a_t, o_{t+1}, r_{t+1})$ throughout the game episode, $e^{TD}$, and populate them in $\mathcal{D}^+$ and $\mathcal{D}^-$ based on a heuristic that depends on—reward, return, and the game states.

First, with LM parameterized by $\theta$ and generating action candidates, we train DRRN for $n^{RL}$ consecutive episodes. After $n^{RL}$ episodes, we sample $d^{LM}$ sized dataset from $\mathcal{D}^+$, and $\mathcal{D}^-$ with probabilities $p^+$ and $1 - p^+$ respectively for 2000 gradient steps at finetuned after every $k$ game steps. To train LM we use a weighted cross-entropy loss:

$$\mathcal{L}^{LM}(\theta) = -\mathbb{E}_{(a_t, o_t) \sim (\mathcal{D}^+, \mathcal{D}^-)} \log P_\theta(a_t \mid o_t) \cdot h\left(\cdot\right) \tag{3.1}$$

Then, we plug-in back the in-game trained LM to recommend actions for the DRRN agent. The maximum buffer size of $\mathcal{D}^+$, $\mathcal{D}^-$, $p^+$, $d^{LM}$, and $n^{RL}$ are all game-specific hyperparameters. The $h\left(\cdot\right)$ is defined as a function of reward $r_t$, or action-advantage, $A(o_t, a_t)$, or assumed 1 uniformly $\forall (o, a) \in \mathcal{O} \times \mathcal{A}$. We evaluate different approaches based on the sampling of transitions, and the loss function ($\mathcal{L}$), used for training the language model. Approaches for LM-in-the-Loop based on the construction of $\mathcal{D}$, and sampling are:

**Un**categorized Transitions (UT): In this setting the transitions stored in the buffer are not categorized by any special heuristic function. We simplify this approach by maintaining a single buffer, $\mathcal{D}$ in place of two. This is a weaker baseline than other heuristics to select useful transitions based on their importance.

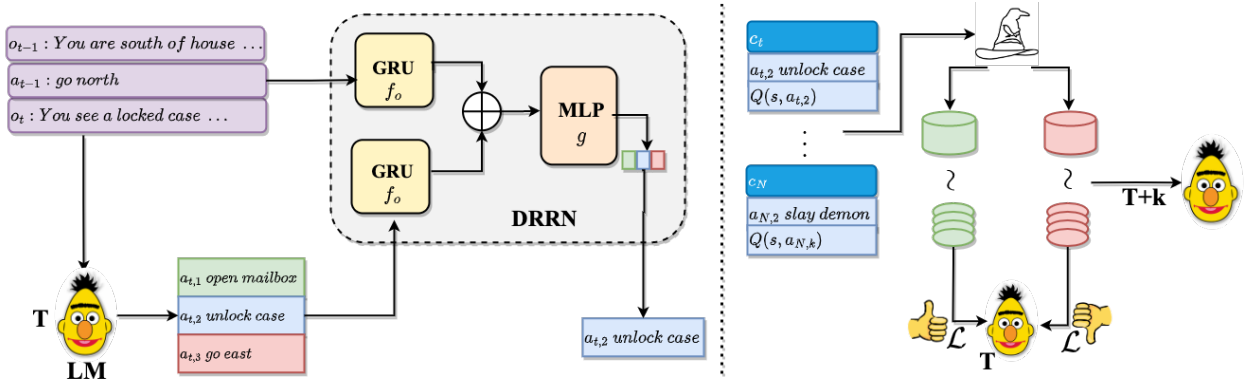

Figure 3.2 Training LM-in-the-Loop post-human-annotated dataset adaptation: RL agent (DRRN) picks the action recommended by the language model (at $T$), which is GPT-2. The context pairs are stored in the replay buffers that are categorized by some heuristic. Then the Language model is updated with in-game transitions after $k$ learning steps in the game. Finally, the updated language model ($T + k$) actions are recommended.

**State Feature Categorized (OC):** In this, the transitions are labeled as useful or not based on whether an action $a_t$ resulted in reward increase or if the agent's location changed. *i.e.,* moved from one room to another. As the location information received is an artifact of the game framework, we consider this as the *Oracle*. Further, we vary $p^+$ to maximize the transitions that encourage exploration to eventually result in improved performance in the game. Here, $h(\cdot)$ is fixed as 1 uniformly $\forall (o, a) \in \mathcal{O} \times \mathcal{A}$.

**Reward Trajectories (RT):** The reward from transitions, $r_t$, is used to categorize positive and negative trajectories. When $r_t > 0$ all transitions up until the earlier non-zero reward are considered positive and added to $\mathcal{D}^+$.

Further, we explore utilizing the return, reward, and advantage function of actions to re-weight $\mathcal{L}^{LM}$ using the $h(\cdot)$ function over **UT** setting as above. We describe them as follows:

**Weighted Cross-Entropy:** In this, the transition data is kept in a single buffer $\mathcal{D}$ similar to in the UT setting. To finetune the language model using the weighted cross-entropy loss (Equation 3.1), we use the exponential weighted advantage function. We use two variants to the weights, wherein UT$^{EA}$ is non-negative using $h(\cdot)$ function:

$$h(o_t, a_t) = e^{\beta \cdot A(o_t, a_t)}, \tag{3.2}$$

where, $\beta \in \mathbb{R}^+$ is a hyperparameter. The other variant, UT$^{LA}$, allows for negative weights with $h(\cdot)$ as follows:

$$h(o_t, a_t) = 1 + \beta \cdot A(o_t, a_t), \tag{3.3}$$

where, $\beta \in \mathbb{R}^+$ is a hyperparameter.

## 3.4 Experiments

We perform comprehensive experiments with LM-in-the-loop set up to study the following questions:

1. Does including the language model in the training loop improve performance?

2. Does LM-in-the-Loop mitigate the reliance on human gameplay transitions?

3. Should the transitions be categorized for improved learning?

### 3.4.1 Task Adaptation Dataset

ClubFloyd dataset [50] is a collection of crawled data from the ClubFloyd website. The dataset comprises of gameplay from experienced players; however, they may not be familiar with the particular games. The data is preprocessed and contains around $217K$ pairs of context an in the form of $((o_{j-1}, a_{j-1}, o_j), a_j)$.

### 3.4.2 Benchmark and the Metric

Jericho [33] is a learning environment that supports human-written interactive fiction games as described in Figure 3.1. We chose 10 games based on the diversity in the challenges faced in each game such as large action space, solution length, and reward sparsity as mentioned in [33,69]. We use the average of the last 100-episodes' score with standard error for individual games [33] as our metric for evaluation.

In addition, we report the average score normalized (avg. norm) against the maximum score possible in each of the games, which estimates the human-machine gap in text-based games. Finally, we also report the relative performance percentage difference between the baseline and the best approach mentioned as $\Delta\%$ in Table 3.1 to capture the improvement as the range of the scores in each game is different.

### 3.4.3  Model Details

Language model (GPT-2) is first finetuned on ClubFloyd dataset. Given the context, $(o_{j-1}, a_{j-1}, o_j)$, the finetuned GPT-2 proposes action candidates for DRRN to choose. Following that, every action candidate and context is encoded with a GRU. Then a decoder combines the representations to estimate the Q-value using a multilayer Perceptron (MLP) and updates the DRRN agent parameter $\Phi$.

During the training process of the DRRN agents, the context-action pairs are stored in the replay buffers. After $k$ steps, we sample $d^{LM}$ sized dataset from $\mathcal{D}^+$, and $\mathcal{D}^-$ with probabilities $p^+$ and $1 - p^+$ respectively and update the language model with in-game transitions. Then, the updated language model is used to propose the action candidates.

The buffer size is defined as $100K$ for replay buffers that uses First-In-First-Out (FIFO) strategy to replace samples. To train, $d^{LM}$ samples are sampled uniformly at random from the two buffers $D^+$ and $D^-$. However, the probability of choosing the buffers are defined by $p^+$ and $p^-$ $(1-p^+)$ respectively. The number of gradient steps for LM training is fixed at 2000 across the set ups. And, across games we experiment with the hyperparameter $p^+ \in [0,1]$ in 0.1 increment, and the value for LM finetuning frequency $k \in [2k, 5k, 10k, 20k]$. The results tabled are estimated from 5 runs.

## 3.5  Results

We follow the questions enumerated in §3.4 to analyze the effect of in-game learning of language models for action recommendations.

### 3.5.1  Effect on Performance

To understand the effect on performance with LM-in-the-Loop, we follow the experimental set up in §3.4.3 to evaluate on Jericho benchmark. Table A.1 compares the different methods detailed in §3.3.2 with reproduced norm score of CALM [50] as the baseline. We see that categorizing the transitions using state features (**OC**) scored the highest in all tasks, suggesting that LM-in-the-Loop enables improved performance. This was also reflected in the avg. norm score with an improvement of $\approx 4\%$ over the baseline. This is $\approx 53\%$ more avg. improvement over the scores obtained by the baseline model. Although the performances of OC are closer to the baseline in many games, the in-game training accelerated the convergence in most games.

However, the improvement with OC is, in a way, a loose upperbound to in-game learning with LM-in-the-Loop, as special techniques to reweight the transitions (UT$^{\text{LA}}$, and UT$^{EA}$), or reward based categorization RT only improved the avg. norm score by $\approx 0.6\%$. On the other hand, the avg. norm score with Uncategorized Transitions (UT) dropped to 19.2% which is $\sim 1\%$ below the baseline performance. The difference in performance between UT, and OC with the baseline suggests that LM-in-the-loop for action recommendation is helpful but requires careful selection of transitions for training the language model.

| Games | CALM | UT | UT$^{\text{LA}}$ | UT$^{\text{EA}}$ | RT | OC | $\Delta(\%)$ | Max Score |
|---|---|---|---|---|---|---|---|---|
| Zork1 | 30.7$_{[4.8]}$ | 32.6$_{[4.4]}$ | 30.4$_{[8.5]}$ | 35.6$_{[5.7]}$ | 30.7$_{[3.8]}$ | **38.0**$_{[1.7]}$ | 23% | 350 |
| Inhumane | 24.8$_{[2.7]}$ | 21.9$_{[5.24]}$ | 28.9$_{[11]}$ | 27.3$_{[3.1]}$ | 29.1$_{[12.7]}$ | **43.4**$_{[3.8]}$ | 75% | 90 |
| Detective | **290.9**$_{[2.7]}$ | 288.5$_{[1.5]}$ | 289.3$_{[0.2]}$ | 288.3$_{[1.3]}$ | 285.1$_{[5.6]}$ | 288.5$_{[1.5]}$ | 0% | 360 |
| Zork3 | 0.3$_{[0.09]}$ | 0.3$_{[0.14]}$ | 0.4$_{[0.1]}$ | 0.6$_{[0.1]}$ | 0.6$_{[0.1]}$ | **0.7**$_{[0.2]}$ | 133% | 7 |
| Omniquest | 6.7$_{[0.3]}$ | 6.0$_{[0.6]}$ | 6.6$_{[0.9]}$ | 6.6$_{[1]}$ | 6.0$_{[0.79]}$ | **7.8**$_{[1.7]}$ | 16% | 50 |
| Library | 11.2$_{[1.3]}$ | 9.3$_{[1]}$ | 9.5$_{[1]}$ | 10.3$_{[0.2]}$ | 10.3$_{[1.8]}$ | **12.1**$_{[0.7]}$ | 8% | 30 |
| Balances | 9.3$_{[0.2]}$ | 9.6$_{[0.1]}$ | 9.6$_{[0.2]}$ | 9.5$_{[0.2]}$ | 9.7$_{[0.2]}$ | **9.7**$_{[0.1]}$ | 4% | 51 |
| Ludicorp | 10.4$_{[0.7]}$ | 11.4$_{[2.6]}$ | 12.5$_{[1.1]}$ | 11.9$_{[2.6]}$ | 11.3$_{[3.1]}$ | **15.1**$_{[0.8]}$ | 45% | 150 |
| Dragon | 0.1$_{[0.06]}$ | 0.1$_{[0.1]}$ | 0.3$_{[0.3]}$ | 0.3$_{[0.3]}$ | 0.1$_{[0.12]}$ | **0.3**$_{[0.2]}$ | 200% | 25 |
| Ztuu | 3.8$_{[0.18]}$ | 4.4$_{[0.0]}$ | 4.5$_{[0.2]}$ | 4.4$_{[0.1]}$ | 4.3$_{[0.1]}$ | **4.5**$_{[0.1]}$ | 18% | 100 |
| Norm Score | 20.1% | 19.1% | 20.6% | 20.9% | 20.7 % | **24.0%** | 52.37% | 100% |

Table 3.1 From the results, it can be consistently seen that LM-in-the-Loop provides a performance improvement over CALM. Especially, categorizing the transitions with state features (OC) scored the highest with $\sim 53\%$ improvement over the scores obtained by the baseline model.

In Figure 3.3, we compare the % of steps in-game learning methods took in average to achieve $k\%$ of CALM model's best performance across the games. We see that LM-in-the-Loop techniques enabled atleast $2\times$ on average acceleration in convergence, although the weaker alternatives to OC with reward based categorization, and reweighted techniques only provided meagre improvements over the baseline (Table 3.1). This shows that the adaptation offered with the ClubFloyd dataset was insufficient, and off-the-shelf techniques can drastically accelerate convergence.

### 3.5.2 Emphasis on Human Annotations

CALM model—the baseline— uses all of the $\sim 220K$ transitions in the ClubFloyd dataset to adapt GPT-2 model for action recommendation. But, by using in-game transitions for LM-in-the-Loop training, the LM is provided with game specific information. So, the requirement for adapting GPT-2 with human annotated transitions should be minimal. [50]

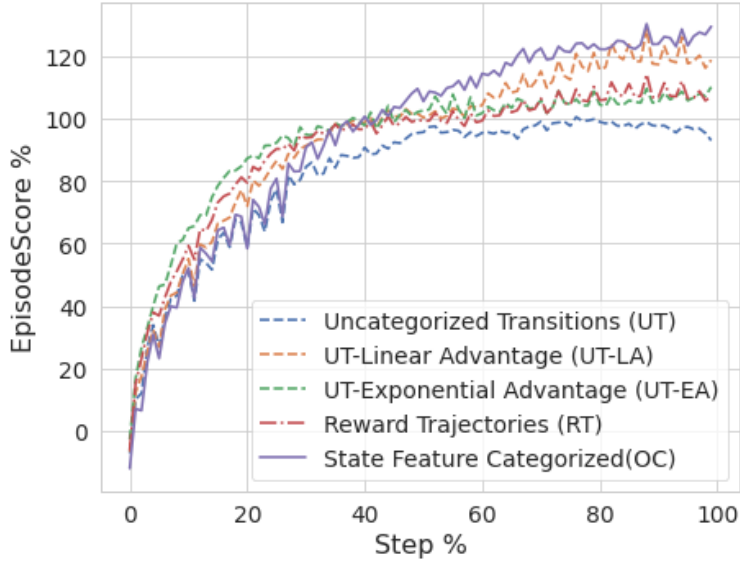

Figure 3.3 We see that LM-in-the-Loop techniques only need half of the steps to achieve the best of CALM. Whereas, using state feature based categorization (OC) achieved better acceleration and performance over the rest.

show that CALM's performance decreased significantly when adaptation was done with 10% of ClubFloyd dataset. The reproduced results of CALM with 10% of adaptation data shows the avg. norm score as 18.5% across the games in Table 3.2. Using State features (**OC**) with 10% of the adaptation date achieved an average norm score of 21.8%, which was more than even using 100% of the adaptation data with CALM. Although there was a small decline in the performance of the detective game, it was insignificant because it was still within the standard error. These results suggest empirically that we can reduce the burden of collecting human-played or human-annotated data by doing in-game learning.

### 3.5.3 Effect of Weight Adjusted LM Loss

Categorization of transitions, although possible in most games, often requires game specific functions to identify what is a good and a bad transition. However, a generalized technique would be to use a notion of the usefulness of transitions that don't require game specific mechanisms. We explore reweighted cross entropy loss as in Equation 3.1 with variations of the $h(\cdot)$ functions from being uniformly distributed as 1 over $(o, a) \in \mathcal{O} \times \mathcal{A}$ to using advantage function with two variations as in Equation 3.2 and Equation 3.3. While UT uses vanilla cross-entropy loss to train the LM on transitions sampled from buffer $\mathcal{D}$, UT$^{\boldsymbol{EA}}$ and UT$^{\boldsymbol{LA}}$ adjusts the experience according to the advantage, $A(o, a)$, of the actions chosen in those observations.

| Games | CALM 100% | CALM 10% | OC 10% |
|---|---|---|---|
| Zork1 | $30.7_{[4.8]}$ | $29_{[3.4]}$ | $\mathbf{35.1}_{[2.3]}$ |
| Inhumane | $24.8_{[2.7]}$ | $15.7_{[14.7]}$ | $\mathbf{27.5}_{[6.8]}$ |
| Detective | $290.9_{[2.7]}$ | $289.5_{[0.2]}$ | $\mathbf{289.6}_{[0.2]}$ |
| Zork3 | $0.3_{[0.09]}$ | $0.6_{[0]}$ | $\mathbf{0.7}_{[0.3]}$ |
| Omniquest | $6.7_{[0.3]}$ | $5.9_{[0.8]}$ | $\mathbf{6.0}_{[1]}$ |
| Library | $11.2_{[1.3]}$ | $\mathbf{10.5}_{[1.5]}$ | $10.2_{[1.8]}$ |
| Balances | $9.3_{[0.2]}$ | $6.6_{[3.5]}$ | $\mathbf{8.6}_{[1.6]}$ |
| Ludicorp | $10.4_{[0.7]}$ | $10.2_{[0.4]}$ | $\mathbf{13.7}_{[0.4]}$ |
| Dragon | $0.1_{[0.06]}$ | $0.1_{[0.06]}$ | $\mathbf{0.3}_{[0.2]}$ |
| Ztuu | $3.8_{[0.18]}$ | $3.6_{[0.1]}$ | $\mathbf{4.1}_{[0.1]}$ |
| Norm | 20.1% | 18.5% | **21.8 %** |

Table 3.2 Using State Features (OC) achieved an average norm score of 21.8% with 10%, which was more than even with CALM using 100% of the adaptation data.

We use causal language modeling to train the GPT-2 LM to discourage the LM in generating a useful action in a state and discouraging the not useful. As $A(o, a) \in [-\infty, +\infty]$, it is important to understand how it affects the language model. A negative advantage for $a'$ in $o'$ should discourage the LM from suggesting $a'$ in $o'$. UT$^{EA}$ re-scales the LM-loss with $h(\cdot) \in [0, 1)$, while UT$^{LA}$ works similar to *Unlikelihood* training as proposed in [70] by maintaining the same scale as $A(o, a)$. But, from the results we see that the differences in reweighting did not tangible affect the performance as seen in Table 3.1 (Columns UT$^{EA}$ and UT$^{LA}$)

## 3.6 Conclusion:

The comparison of LM-in-the-Loop with baseline and their absolute performances from Table 3.1 shows that there is more room for improvement. Despite the LMs having strong linguistic priors from pretraining, the large action space when it comes to generative task is one of the significant challenges in adapting LMs to text-based games. Although interactive learning is promising, towards realizing interactive task solving agents, it is imperative to address the issues due to scalability and data-efficiency. The results in the paper through exploring the possibility of adapting language models for action suggestions through utilizing the in-game generated transitions opens up discussions on several key questions:

While there is improvement in performance, and acceleration in comparison to not learning from the game transitions, the absolute improvement with respect to the games has still a long way to go. When DRRN module was plugged out for ablation, the argmax action of LM was not even close to a reasonable performance indicating the heavy lifting in planning was from DRRN. Towards realizing LMs in situated learning environments, adapting LMs to different games is a challenging language understanding milestone. Specifically, it is important to align LM's action generation likelihood to reflect the action value function.

Despite the acceleration and a reduced need for human transitions to adapt LMs for action suggestion, interpreting their performance through the conventional lens of automatic semantic and syntax scores is less effective. It is, then, only imperative to make the application of LMs in text games interpretable through automatic metrics that identifies important transitions to train LM-in-the-Loop.

## Limitations

The paper analyzes the possibility and challenges in LM-in-the-Loop training of GPT-2 model for action recommendation in text based games. The claims in the work can be further supported with experiments on different LLM. Similarly, the generalization experiments could have added more support to the lack of evidence with additional games. However, these are compute intensive experiments and the claims are largely made in consideration to the limitations in the set up.

## Acknowledgements

Sarath Chandar is supported by a Canada CIFAR AI Chair and an NSERC Discovery Grant. The authors acknowledge the computational resources provided by the Digital Research Alliance of Canada and Mila Compute resources. We are thankful to Siva Reddy for their helpful feedback in this work.

# CHAPTER 4    Language Model as Multi-Agent

This chapter is a slightly modified version of our paper [71].

**My contributions as First Author:** Problem statement identification, literature survey, benchmark establishment, implementation and running experiments.

**Status:** The paper is currently accepted at ICLR 2024 - Generative Models for Decision Making Workshop under the title "Multi-agent text-based Hanabi challenge" [71, 72].

## 4.1   Introduction

In both daily life and work, humans are constantly involved in coordination tasks, whether it's navigating public transportation, organizing events, or chatting with language model to accomplish a task. These diverse activities necessitate individuals to synchronize their efforts and engage in collaborative decision-making to achieve successful outcomes [73–75]. Modern LLMs such as ChatGPT [76], LLaMA [32], Mistral [77], showcase remarkable zero-shot or few-shot generalization capacities, particularly in complex natural language tasks. As these models become more widely adopted for supporting humans in monotonous and arduous tasks, the necessity to effectively collaborate and coordinate with humans or other autonomous systems in task completion becomes essential.

Hanabi [78] game serves as a rigorous benchmark for evaluating the coordination and co-operation for MARL agents. Hanabi, a popular cooperative game designed for two to five players, with approximately $2^{90}$ unique player hands in a five-player setting. Progressing in the game requires intricate skills, including long-term planning, adaptability to new partners (cross-play), precise assistance through clues to other agents, and complex reasoning. Adding to the complexity, players are required to infer the beliefs and intentions of their counterparts through theory of mind reasoning. In our work, we propose a novel Hanabi-text challenge that serves as a testbed for the current large language model to operate on a textual environment in a multi-agent setup.

Our contributions of this work is as follows,

- We propose a novel text version of the Hanabi game which can be a pathway to use language model acting in a multi-agent setting.

- We show that a language model finetuned with expert trajectory can act independently as an agent in the Hanabi-text environment.

- We have successfully integrated a finetuned language model with a Q-learning RL agent via distillation loss to improve performance and sample efficiency.

## 4.2 Related Work

**Reinforcement learning for text-based games:** Inspired by the remarkable success of deep reinforcement learning (RL) techniques in diverse fields such as gaming [79], robotics [80,81], and natural language processing [82], there has been a surge of interest in developing game agents based on RL for text-based games. In such environments [33,83–86], the agent is presented with a textual description of a goal and must take actions to achieve the objective [87]. These interactive environments offer challenging and realistic training that requires a solid understanding of the language and the task. Moreover, connecting language with the physical world is critical to solving the task [47,88]. To address these, researchers have developed several RL-based agents that operate on text [83,89–91]. Language models have been used to propose action candidates [42,92–94]. Most of the current works focus on the single-agent setup, however ours is a multi-agent text environment. [95–97]: reasoning about other agents with their own mental states. There are some works [20,98] on multi-agents that operate on text in an open world. However, evaluation of generative agents in that setup would be a major challenge.

**Hanabi:** [12] proposed the Hanabi challenge as a new frontier for AI research, presenting unique multi-agent learning challenges due to its combination of purely cooperative gameplay and imperfect information. There have been several methods trying to solve the game via self-play (SP) [99,100] however, fall short to coordinate with other novel agents. This highlighted the necessity to train MARL agents with generalization capabilities. Subsequently, the concept of Zero-Shot Coordination (ZSC) was introduced by [18], emphasizing the ability to cooperate with new agents. Following this, the research community has concentrated on devising methods to enable pre-trained RL agents to achieve Zero-Shot Coordination (ZSC) with novel, unseen agents, as evidenced by [101–105], Few-shot Coordination [106]. [107] proposed to use pretrained large language models to generate a prior policy conditioned on the human instruction to regularize the RL objective. However, this prior policy is only conditioned on other agents' actions and heuristic instructions in contrast to our work where the LM operates on both observations and action avoiding the need to design hand-coded instructions.

## 4.3 Methodology

### 4.3.1 Understanding the Knowledge Representations of LM Agents for Hanabi

To train LLMs to play Hanabi, we need to convert the state description provided by the game environment to the text format understandable by the LLMs. For this, we use a template as shown in Figure 4.1a. Using this template, we started to collect a dataset from a pre-trained expert Hanabi agents using Off-belief Learning [103], SOTA ZSC method. We refer to appendix B for more details on the dataset curation.

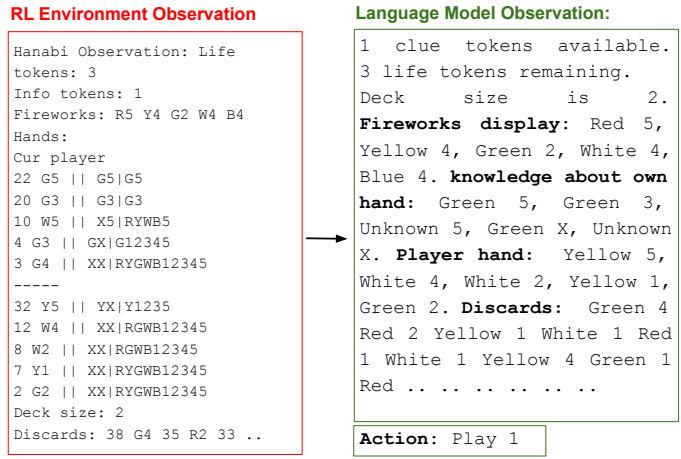

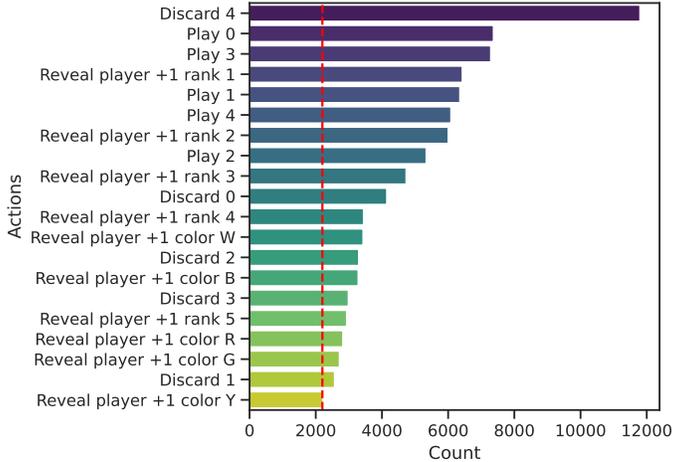

Figure 4.1 *a) Converting Hanabi RL environment observation to text to be processed by the LMs. b) Visualizing the number of actions available in the dataset to create a diverse dataset of Hanabi gameplay in the form of text.*

### 4.3.2 Dataset details

The dataset is acquired through self-play mode, utilizing a pre-trained OBL agent in the Hanabi game. Trajectories are filtered selectively with a gameplay score exceeding 20. Then, these trajectories are broken down into state-action pairs to suit language model training. During the initial data exploration, we found the action categories are imbalanced as shown in 4.1b, hence the language model overfits to discard 4 based on the confusion matrix for the prediction. To avoid that, we did categorical sampling consisting of 2200 samples per action type, aggregating to $44,000$ instances. Then we checked for duplicate states and dropped them, there were approximately 100 duplicates as this could mislead the model's learning. After which, 10% of the dataset is reserved for testing by random sampling. Further, the dataset is split into 90% for train and 10% for validation.

### 4.3.3 Supervised Finetuning of Language Models

This work focuses on the implementation of two categories of language models such as classifier and generative models to act as an agent in the RL environment. To finetune our language models, we minimize the cross-entropy loss between the true and predicted action labels. The language models used in our work are as follows:

**BERT** [25] is a powerful transformer-based model designed for natural language processing tasks. It is pre-trained on large corpora and captures bidirectional context information, allowing it to understand context-dependent meanings of words in a sentence.

**DistilBERT** [108] is a distilled version of BERT, designed to retain most of its performance while being computationally more efficient. It achieves this by reducing the number of parameters and layers, making it a suitable candidate model for integrating with RL.

**GPT-2 Classifier** [43], we add a linear layer on top of pre-trained layers of the decoders model with its output dimension equaling our number of labels.

**GPT-2 Generative** [43] language model known for its capacity to generate coherent and relevant text which acts as a generator.

### 4.3.4 Integrating LM Hanabi in the RL loop

One of the applications of having a pretrained language agent is to improve the transfer and convergence between RL agents. Inspired by [16], we distill the knowledge of a language agent (Teacher policy) to the RL agent (Student policy) that is learning from scratch. In equation 4.1, the goal of the RL agent is to minimize its own TD loss [109], and a distillation loss estimated using the language model as the teacher policy ($\pi_{LM}$). The expectation, denoted by $\mathbb{E}_{s \sim \mathcal{D}}$ is taken over the states $s$ sampled from the replay buffer.

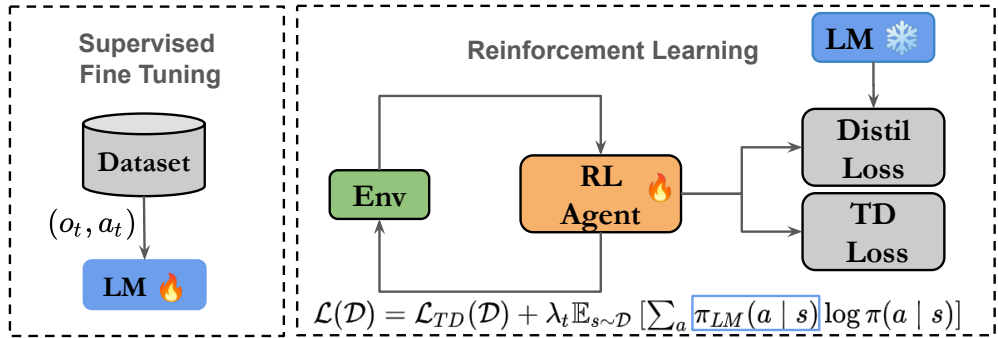

Figure 4.2 *Integration of LM into RL loop. First, we finetune an LM using an expert dataset. Then, we distill the pre-trained LM knowledge into the RL policy using distillation loss while RL is interacting with the environment.*

$$\mathcal{L}(\mathcal{D}) = \mathcal{L}_{TD}(\mathcal{D}) + \lambda_t \mathbb{E}_{s \sim \mathcal{D}} \left[ \sum_a \pi_{LM}(a \mid s) \log \pi(a \mid s) \right] \quad (4.1)$$

Where, $\pi_{LM}(a \mid s)$ represents the predicted probability of taking action $a$ a given state $s$ according to the teacher agent - language model. $\log \pi(a \mid s)$ signifies the log likelihood of the action $a$ based on the student policy. $\lambda_t$ acts as a regularization, which is a hyperparameter for controlling the importance of the teacher policy.

## 4.4 Experiments

### 4.4.1 Language model as Agents

As a first experiment, we apply low-rank finetuning on LLaMA-7B [110] with different rank and small dataset size to align it better to the Hanabi game distribution. However, it performs poorly based on the gameplay score. [107] further affirms that current LLMs are still far from being capable of solving Hanabi independently.

We further explore full funetuning of language model variants like BERT, DistilBERT, GPT-2 Classifier, and GPT2-Generative that can be integrated within the RL loop effectively. We selected the optimal checkpoint of each language model based on gameplay scores. The best checkpoint is then subjected to 1200 runs in the Hanabi environment to handle variance/randomness, as depicted in Figure 4.3a. Both the BERT and DistilBERT models demonstrate a commendable performance in the Hanabi gameplay, achieving a maximum score of 23 out of a possible 25. Their average game play scores hover around 10 during the gameplay. The GPT2-generative model has better top-k test accuracy, however, it fails short compared to the classification-based model in the overall gameplay score with $\sim 4.5$. We further tried with different percentages of training datasets to understand the role of data. Compared to 10% or less, when using 25% of the data there is a sharp increase in the gameplay score. However, the performance plateaus for both 75% and 100%, are indicative of reaching a saturation point. Also, we tried different BERT variants, and all are saturated to the same game score irrespective of the increase in the parameter size. Finally, we also investigated the role of discarding information in the observation and found it didn't help much in the gameplay score.

### 4.4.2 Understanding the Knowledge Representations of LM Agents for Hanabi

To investigate knowledge possessed by the language model about the Hanabi game we evaluate the top-k prediction accuracy of four language models. Employing a comprehensive test set comprising 4400 instances, we calculate the top-k accuracy for varying values of $k$, ranging from 1 to 5. Our objective is to gauge the models' efficacy within the top-k predictions of the expert agent's data. When $k = 1$(greedy action), the test accuracy is around $\sim 54\%$, however, when k=5 more than 90% of the ground truth actions are among the top-5 predictions of the language model. As the $k$ increases, the test accuracy increases which shows a clear pattern that the optimal actions exist in the top-k predictions as shown in Figure 4.3b. On top of that, we check whether the predicted actions are within legal actions. Even though the legal actions are not provided explicitly in the observation, the legal actions are learned implicitly by these models. For BERT, DistilBERT, and GPT-2 generative model has 100% for $k = 1$ and there is a slight drop as we increase the $k$ value as shown in Figure 4.3c. This motivates us that, the language models contain valuable information about the overall dynamics of the game and how to play the game strategically.

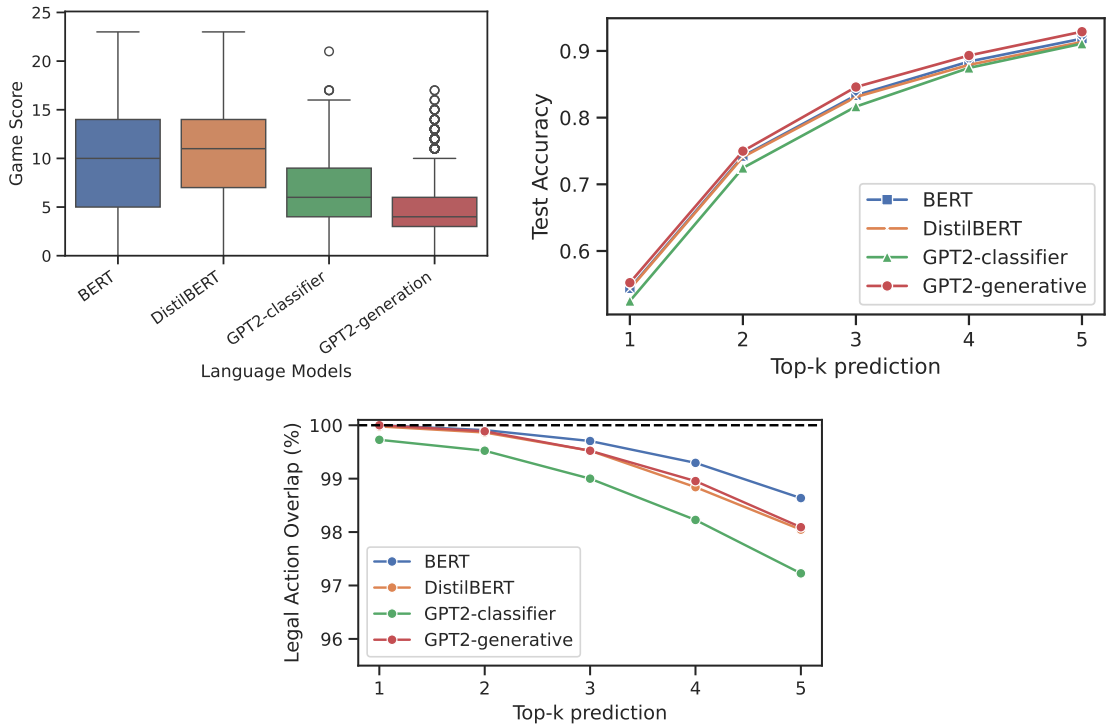

Figure 4.3 *The evaluation of language models performance as an agent is conducted through three key metrics: a) gameplay score, measured across various language models with 1200 game runs, b) top-k predictions to assess the prediction label alignment with test labels, and c) top-k predictions to check the overlap to legal actions.*

### 4.4.3 Integrating LMs to RL-loop

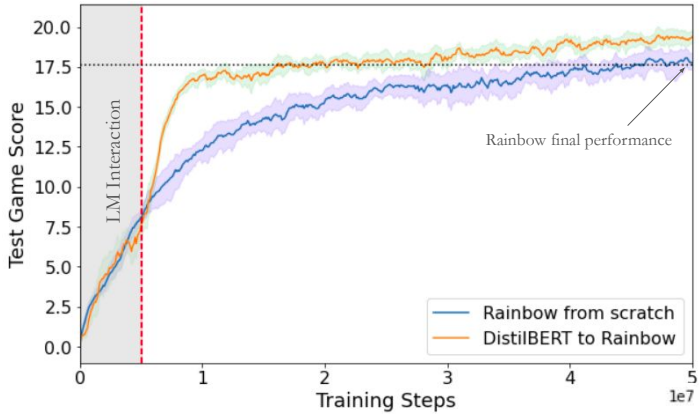

Figure 4.4 *Rainbow agent integrated with distillation loss. The Red dashed line represents the timestep that we stop interacting with the language model. The black dotted line represents the final performance of the rainbow agent trained from scratch.*

We integrate the best model (DistilBERT) from sec 4.4.1 in terms of game-play to the training loop of a Rainbow agent [111] using Equation 4.1. For the first 5m steps, $\lambda$ is set to 1 meaning that the RL policy is only learning from the distillation loss. After that, the Rainbow agent is updated using the standard TD loss. We use the scheduler to wean off the student's dependency on the teacher policy as training progresses. As shown in Figure 4.4, integrating the distillation loss results in a jump in the performance and becoming almost 5x more sample efficient.

## 4.5   Conclusion

In this work, we proposed a text-based version of the Hanabi game. This is a challenging testbed for the current LLMs due to the need for long-term planning, adaptability to new partners (cross-play), precise assistance through clues to other agents, complex reasoning, and theory of mind. We trained a language model using expert trajectories and deployed it as an agent in this testbed, achieving a notable game score, yet it remains far from being fully solved. Furthermore, we show that language model's knowledge can be distilled into an RL agent learning from scratch. In conclusion, text-based Hanabi can pave the way for the NLP community to utilize this Multi-agent testbed to further test and advance the capabilities of the current LLMs.

## Acknowledgements

Sarath Chandar is supported by a Canada CIFAR AI Chair and an NSERC Discovery Grant. The authors acknowledge the computational resources provided by the Digital Research Alliance of Canada and Mila Compute resources. We are thankful to Xutong Zhao, Mathieu Raymond, Kamran Chitsaz, Quentin Fournier for their helpful feedback in this work.

# CHAPTER 5    FUTURE WORK

Our findings in Chapters 3 and 4 on LLMs are a step towards: How can we design generalist multi-agents that combine language models to transfer across different players and zero/few-shot adaptation to novel partners? As humans can utilize knowledge between Hanabi games with varying numbers of players, the underlying principle of the game remains the same. This proposal seeks to address this challenge based on two key observations. First, humans can better **coordinate** with a policy if it can concisely be summarized in natural language. Second, in most real-world coordination scenarios, humans talk to each other or even negotiate to achieve some agreement on how they should **collaborate**. The MARL community explores these two critical aspects, and we aim to incorporate them to guide agents toward more human-like policies.

We aim to devise algorithms, benchmarks, and frameworks that empower agents to synchronize their actions, communicate effectively, and collaborate towards shared objectives by taking inspiration from MARL to build Multi-Agent Language Models.

Our goal is to determine and improve upon the reasoning ability of LLMs to solve multi-agent coordination tasks. Effective transfer learning is essential as the tabula rasa approach is impractical in large-scale systems, which is a significant bottleneck in training generalist agents. Finally, agents must be capable of adjusting to unfamiliar partners and demonstrate robustness to partners.

Goal 1: *How to benchmark theory of mind capabilities of current LLMs?*

Goal 2: *How can we knowledge transfer among games with different number of agents?*

Goal 3: *How can we leverage the LLM's few-shot reasoning capabilities to quickly adapt to novel partners at test time?*

In pursuit of goals 1, 2, and 3, we have investigated into the following crucial problem statements essential to this proposal.

In Chapter 3, we present empirical evidence that despite the strong linguistic priors of LMs from pretraining, the large action space in generative tasks poses a significant challenge in adapting LMs to text-based games. While interactive learning shows promise, the road to realizing interactive task-solving agents is paved with scalability and data efficiency challenges. Our research delved into these complexities, exploring the possibility of adapting language models for action suggestions through in-game-generated transitions.

These insights are crucial as adaptive action spaces are necessary for Goal 2 for transfer learning and minimizing human annotations necessary for Goals 2 and 3.

In Chapter 4, we trained a language model using expert trajectories. We deployed it as an agent in a novel-text-based Hanabi testbed, achieving a notable game score, yet it remains far from being fully solved. Furthermore, we show that language model's knowledge can be distilled into an RL agent learning from scratch. Text-based Hanabi can pave the way for the Natural Language Processing community to utilize this Multi-agent testbed to test further and advance the capabilities of the current LLMs.

This insight is helpful towards Goal 1 as text-based Hanabi is necessary to investigate the theory of mind for LLMs. Furthermore, textual representation aids in building robust models that help in transfer learning (Goal 2) and few-shot adaptation (Goal 3), a step towards generalist agents.

Based on the insights from our previous works, the following are the future directions:

## 5.1   Hanabi-Text Environment for Multi Agent LM:

To establish and understand the theory of mind of current LLMs, we will train RL agents through self-play (e.g., Rainbow, PPO) on Hanabi, from 2 players to 5 players. Each RL agent will have learned a specific strategy. Next, we will test cross-play by pairing these agents with LLM agents with different architectures and model sizes (e.g., GPT-3 with 1B, 3B, or 7B parameters, Llama 7B, 13B, 70B). We will perform experiments with an increasing number of LLM agents. Moreover, we will allow varying degrees of fine-tuning by allowing the LLM agents to review a number of playthroughs performed solely by RL agents.

By providing systematic benchmarks across different model architectures and sizes, our research has the potential to significantly advance the understanding of how well LLMs can learn RL strategies. It will also shed light on how well they will interact with other non-adaptive agents and how well they will collaborate. Given the complexity of the game and the limited amount of observed playthroughs, they will have to infer general patterns, a challenge that our research aims to address. This will also guide us in determining the most appropriate model type that will serve as a starting point for our other research questions, inspiring further exploration in this field.

## 5.2 DRRN-based LLMs:

We aim to incorporate a language model to enable knowledge transfer, a step toward a generalist agent. In the existing approach, a DQN network [37] is used, which takes in the state information and predicts the action. However, it needs to be more robust to change in state/action space and dynamic action space when moving from 2 players to a 3/4/5 player game. Hence, we aim to use a DRRN architecture [41] with language model, which is used predominantly in text-based games as it is robust and can handle dynamic action space. We propose a novel architecture incorporating a trained BERT model [25] with DRRN with updatable parameters based on the loss function from MLP at the top to predict the Q-values. It enables the transfer of prior knowledge and helps accelerate and improve performance to learn more informed policies.

## 5.3 Few-shot coordination to novel partners

Towards adaptation to novel partners, [107] proposed a technique for training RL agents to follow instructions based on natural language that specify how a human wishes to communicate with AI to facilitate human-AI coordination. RL is guided to converge to the most desirable equilibria by using a prior policy created by LLMs based on natural language instruction instead of collecting labeled human data. However, it requires a closed model like ChatGPT and it doesn't take into the observation space into account while estimating the prediction probability.

We aim to have open-source models that can learn to operate on the observational space of the Hanabi game. Also, our LLM-based agent encodes the agent's policy based on context and the general knowledge accumulated during pre-training; our agent will have a separate action head for each partner. This head will be trained to predict the partner's actions. After this, in the finetuning phase, it can be adapted with a few game trajectories.

As an evaluation procedure, we will make teams of LLM-based agents that have been finetuned on different strategies. We will then analyze how well they learn the strategy of others by comparing each other's action heads and how well they converge toward a common strategy. Moreover, we will create an evaluation setup where a new LLM agent is added or removed, thus changing the number of players in the game. Coping with changes of players in Hanabi has yet to be explored, which would greatly benefit the community, as this is a common occurrence in everyday life. We believe that, through language, our LLM agents will be able to cope with such a change.

# CHAPTER 6    RISKS

The ability to cooperate with other agents and humans is a desirable characteristic of an interactive agent that learns in the presence of other agents and is a fundamental capability to thrive as a society. Even though the different research questions are expressed chronologically, they are independent of the successful completion of the previous work package before we can start investigating them. Goal 2 and Goal 3 can be performed independently. This allows us to make efficient use of time and resources, as we do not have to wait for the results of one research question before tackling another.

Nonetheless, it is also important to acknowledge several risks involved in this process:

- When language models are finetuned towards reward maximization on a task, they gradually lose language's syntactic and semantic properties. This phenomenon is called language drift, which could result in specialization for specific tasks rather than effective generalization to handle unseen tasks. This problem can be mitigated by employing low-rank adaptation, freezing certain layers of the language model selectively, or using the divergence penalty.

- Hanabi game with RL/LLM could raise the source of errors through randomness. Especially in Hanabi gameplay, even with 1000 gameplays, the variance is high due to the stochasticity of the game based on the randomness in card distribution, card deck, start of the player, and player strategy. Hence, to obtain more reliable results, we plan to run for more seeds than the typical RL benchmark, and increased gameplay for a statistically significant result.

- LLMs often have slower inference speeds, and there are significant engineering problems when combining them with RL with millions of interactions. Small-sized language models with improved representation capabilities can mitigate these problems and produce performance on par with bigger models.

# CHAPTER 7    TIMELINE

|  | 2022 | 2023 | | | 2024 | | | 2025 | | | 2026 | |
|---|---|---|---|---|---|---|---|---|---|---|---|---|
|  | F | W | S | F | **W** | S | F | W | S | F | W | S |
| **PhD Candidate** |  |  |  |  |  |  |  |  |  |  |  |  |
| Course Work | ■ | ■ | ■ |  |  |  |  |  |  |  |  |  |
| Comprehensive exam |  |  |  | ■ |  |  |  |  |  |  |  |  |
| Proposal defense |  |  |  |  | ■ |  |  |  |  |  |  |  |
| Thesis writing |  |  |  |  |  |  |  |  |  |  | ■ |  |
| Thesis defense |  |  |  |  |  |  |  |  |  |  |  | ■ |
| **LM in the loop:Towards Adaptation - Chapter 3** |  |  |  |  |  |  |  |  |  |  |  |  |
| Motivate problem | ■ |  |  |  |  |  |  |  |  |  |  |  |
| Literature Survey | ■ |  |  |  |  |  |  |  |  |  |  |  |
| Propose method |  | ■ |  |  |  |  |  |  |  |  |  |  |
| Write code Run Experiment |  |  | ■ |  |  |  |  |  |  |  |  |  |
| Write paper |  |  |  | ■ |  |  |  |  |  |  |  |  |
| **LM as multi-agent - Chapter 4** |  |  |  |  |  |  |  |  |  |  |  |  |
| Motivate problem |  |  |  | ■ |  |  |  |  |  |  |  |  |
| Literature survey |  |  |  | ■ |  |  |  |  |  |  |  |  |
| Propose method |  |  |  |  | ■ |  |  |  |  |  |  |  |
| Write code Run experiment |  |  |  |  | ■ |  |  |  |  |  |  |  |
| Write paper |  |  |  |  | ■ |  |  |  |  |  |  |  |
| **Hanabi-Text    Environment    for Multi Agent LM** |  |  |  |  |  |  |  |  |  |  |  |  |
| Motivate problem |  |  |  |  |  | ■ |  |  |  |  |  |  |
| Literature survey |  |  |  |  |  | ■ |  |  |  |  |  |  |
| Write code run experiment |  |  |  |  |  |  | ■ |  |  |  |  |  |
| Write paper |  |  |  |  |  |  |  | ■ |  |  |  |  |
| **DRRN-based LLMs:** |  |  |  |  |  |  |  |  |  |  |  |  |
| Motivate problem |  |  |  |  |  |  |  | ■ |  |  |  |  |
| Literature survey |  |  |  |  |  |  |  | ■ |  |  |  |  |
| Write code run experiment |  |  |  |  |  |  |  |  | ■ |  |  |  |
| Write paper |  |  |  |  |  |  |  |  |  | ■ |  |  |
| **Few-shot    coordination    to    novel partners** |  |  |  |  |  |  |  |  |  |  |  |  |
| Motivate problem |  |  |  |  |  |  |  | ■ |  |  |  |  |
| Literature survey |  |  |  |  |  |  |  | ■ |  |  |  |  |
| Propose method |  |  |  |  |  |  |  |  | ■ |  |  |  |
| Write code Run experiment |  |  |  |  |  |  |  |  |  |  | ■ |  |
| Write paper |  |  |  |  |  |  |  |  |  |  | ■ |  |
| **Teaching** |  |  |  |  |  |  |  |  |  |  |  |  |
| Teaching Assistantship |  |  |  |  |  |  | ■ | ■ |  |  |  |  |

Table 7.1 Milestone Timeline

# CHAPTER 8    CONCLUSION

As LLMs are everywhere and will continue to spread, and we increasingly interact with them, effective collaboration and coordination with humans or other autonomous systems become imperative. Understanding the meaning behind other people's actions and being able to infer what they will do in the future is key towards collaborative success. In chapter 3, We present empirical evidence that despite the strong linguistic priors of LMs from pretraining, the large action space in generative tasks poses a significant challenge in adapting LMs to text-based games. We also have explored a language model in a single-agent setup that can adapt to in-game trajectories to reduce the reliance on human-annotated data.

Following this, in chapter 4, we investigated language models in a multi-agent setup through the novel Hanabi-text environment, which serves as a unique test bed for building multi-agent language models. We trained a language model using expert trajectories, showed successful distillation, and deployed it as an agent in this testbed, achieving a notable game score. Yet, it remains far from being fully solved.

Using the insights from our works, our goal is to explore further the following: 1) Benchmarking theory of mind capabilities of current LLMs to assess how well LLMs can understand and participate in cooperative play with other LLMs by interpreting their intentions. 2) In large-scale RL systems, the tabula rasa approach is impractical and a significant bottleneck in training generalist agents. Hence, exploring knowledge transfer approaches among Hanabi games with different numbers of agents that accelerate training and increase performance becomes essential. 3) Finally, agents using different learning algorithms will converge to various strategies and make different assumptions about what their partners will do. When paired together, these agents will underperform. How can we leverage the LLM's few-shot reasoning capabilities to adapt to novel partners quickly during test time?

Through this proposal, we aim to significantly advance the adaptation capabilities of artificial agents towards others, both when paired with other artificial agents or humans. We push towards this goal through LLM-based agents, with whom we already regularly interact, and which are capable of producing natural language. As LLMs are everywhere and will continue to spread, and we increasingly interact with them, the outcomes of this proposal have the potential to significantly impact society, creating truly hybrid systems.

# REFERENCES

[1] R. S. Sutton and A. G. Barto, *Reinforcement Learning: An Introduction*. Cambridge, MA, USA: A Bradford Book, 2018.

[2] OpenAI, "GPT-4 Technical Report," *arXiv e-prints*, p. arXiv:2303.08774, Mar. 2023.

[3] Gemini Team, "Gemini: A Family of Highly Capable Multimodal Models," *arXiv e-prints*, p. arXiv:2312.11805, Dec. 2023.

[4] Anthropic, "The Claude 3 Model Family: Opus, Sonnet, Haiku," *arXiv e-prints*, 2024.

[5] T. Kojima, S. S. Gu, M. Reid, Y. Matsuo, and Y. Iwasawa, "Large language models are zero-shot reasoners," *Advances in neural information processing systems*, vol. 35, pp. 22 199–22 213, 2022.

[6] D. Vilar, M. Freitag, C. Cherry, J. Luo, V. Ratnakar, and G. Foster, "Prompting palm for translation: Assessing strategies and performance," *arXiv preprint arXiv:2211.09102*, 2022.

[7] J. Zhang, Y. Zhao, M. Saleh, and P. Liu, "Pegasus: Pre-training with extracted gap-sentences for abstractive summarization," in *International conference on machine learning*. PMLR, 2020, pp. 11 328–11 339.

[8] N. Chomsky, "On the nature of language," *Origins and evolution of language and speech*, vol. 280, pp. 46–57, 1976.

[9] L. Steels, "The synthetic modeling of language origins," *Evolution of communication*, vol. 1, no. 1, pp. 1–34, 1997.

[10] A. Budanitsky and G. Hirst, "Evaluating wordnet-based measures of lexical semantic relatedness," *Computational linguistics*, vol. 32, no. 1, pp. 13–47, 2006.

[11] C. D. Manning, "Computational linguistics and deep learning," *Computational Linguistics*, vol. 41, no. 4, pp. 701–707, 2015.

[12] N. Bard, J. N. Foerster, S. Chandar, N. Burch, M. Lanctot, H. F. Song, E. Parisotto, V. Dumoulin, S. Moitra, E. Hughes *et al.*, "The hanabi challenge: A new frontier for ai research," *Artificial Intelligence*, vol. 280, p. 103216, 2020.

[13] D. Premack and G. Woodruff, "Does the chimpanzee have a theory of mind?" *Behavioral and brain sciences*, vol. 1, no. 4, pp. 515–526, 1978.

[14] N. Rabinowitz, F. Perbet, F. Song, C. Zhang, S. A. Eslami, and M. Botvinick, "Machine theory of mind," in *International conference on machine learning*. PMLR, 2018, pp. 4218–4227.

[15] H. Touvron, L. Martin, K. Stone, P. Albert, A. Almahairi, Y. Babaei, N. Bashlykov, S. Batra, P. Bhargava, S. Bhosale, D. Bikel, L. Blecher, C. Canton Ferrer, M. Chen, G. Cucurull, D. Esiobu, J. Fernandes, J. Fu, W. Fu, B. Fuller, C. Gao, V. Goswami, N. Goyal, A. Hartshorn, S. Hosseini, R. Hou, H. Inan, M. Kardas, V. Kerkez, M. Khabsa, I. Kloumann, A. Korenev, P. Singh Koura, M.-A. Lachaux, T. Lavril, J. Lee, D. Liskovich, Y. Lu, Y. Mao, X. Martinet, T. Mihaylov, P. Mishra, I. Molybog, Y. Nie, A. Poulton, J. Reizenstein, R. Rungta, K. Saladi, A. Schelten, R. Silva, E. M. Smith, R. Subramanian, X. E. Tan, B. Tang, R. Taylor, A. Williams, J. X. Kuan, P. Xu, Z. Yan, I. Zarov, Y. Zhang, A. Fan, M. Kambadur, S. Narang, A. Rodriguez, R. Stojnic, S. Edunov, and T. Scialom, "Llama 2: Open Foundation and Fine-Tuned Chat Models," *arXiv e-prints*, p. arXiv:2307.09288, Jul. 2023.

[16] R. Agarwal, M. Schwarzer, P. S. Castro, A. C. Courville, and M. Bellemare, "Reincarnating reinforcement learning: Reusing prior computation to accelerate progress," *Advances in Neural Information Processing Systems*, vol. 35, pp. 28 955–28 971, 2022.

[17] H. Nekoei, X. Zhao, J. Rajendran, M. Liu, and S. Chandar, "Towards few-shot coordination: Revisiting ad-hoc teamplay challenge in the game of hanabi," in *Proceedings of The 2nd Conference on Lifelong Learning Agents*, ser. Proceedings of Machine Learning Research, S. Chandar, R. Pascanu, H. Sedghi, and D. Precup, Eds., vol. 232. PMLR, 22–25 Aug 2023, pp. 861–877. [Online]. Available: https://proceedings.mlr.press/v232/nekoei23b.html

[18] H. Hu, A. Lerer, A. Peysakhovich, and J. Foerster, ""Other-play" for zero-shot coordination," in *Proceedings of the 37th International Conference on Machine Learning*, ser. Proceedings of Machine Learning Research, H. D. III and A. Singh, Eds., vol. 119. PMLR, 13–18 Jul 2020, pp. 4399–4410.

[19] S. Agashe, Y. Fan, A. Reyna, and X. E. Wang, "Llm-coordination: Evaluating and analyzing multi-agent coordination abilities in large language models," 2024.

[20] J. S. Park, J. C. O'Brien, C. J. Cai, M. R. Morris, P. Liang, and M. S. Bernstein, "Generative agents: Interactive simulacra of human behavior," 2023.

[21] W. X. Zhao, K. Zhou, J. Li, T. Tang, X. Wang, Y. Hou, Y. Min, B. Zhang, J. Zhang, Z. Dong, Y. Du, C. Yang, Y. Chen, Z. Chen, J. Jiang, R. Ren, Y. Li, X. Tang, Z. Liu, P. Liu, J.-Y. Nie, and J.-R. Wen, "A survey of large language models," 2023.

[22] Q. Fournier, G. M. Caron, and D. Aloise, "A practical survey on faster and lighter transformers," *ACM Comput. Surv.*, vol. 55, no. 14s, jul 2023. [Online]. Available: https://doi.org/10.1145/3586074

[23] A. Vaswani, N. Shazeer, N. Parmar, J. Uszkoreit, L. Jones, A. N. Gomez, L. Kaiser, and I. Polosukhin, "Attention is all you need," *CoRR*, vol. abs/1706.03762, 2017. [Online]. Available: http://arxiv.org/abs/1706.03762

[24] J. Chorowski, D. Bahdanau, D. Serdyuk, K. Cho, and Y. Bengio, "Attention-based models for speech recognition," *CoRR*, vol. abs/1506.07503, 2015. [Online]. Available: http://arxiv.org/abs/1506.07503

[25] J. Devlin, M.-W. Chang, K. Lee, and K. Toutanova, "BERT: Pre-training of deep bidirectional transformers for language understanding," in *Proceedings of the 2019 Conference of the North American Chapter of the Association for Computational Linguistics: Human Language Technologies, Volume 1 (Long and Short Papers)*, J. Burstein, C. Doran, and T. Solorio, Eds. Minneapolis, Minnesota: Association for Computational Linguistics, Jun. 2019, pp. 4171–4186. [Online]. Available: https://aclanthology.org/N19-1423

[26] X. Zhang, J. Zhao, and Y. LeCun, "Character-level convolutional networks for text classification," in *Advances in Neural Information Processing Systems*, C. Cortes, N. Lawrence, D. Lee, M. Sugiyama, and R. Garnett, Eds., vol. 28. Curran Associates, Inc., 2015. [Online]. Available: https://proceedings.neurips.cc/paper/2015/file/250cf8b51c773f3f8dc8b4be867a9a02-Paper.pdf

[27] A. L. Maas, R. E. Daly, P. T. Pham, D. Huang, A. Y. Ng, and C. Potts, "Learning word vectors for sentiment analysis," in *Proceedings of the 49th Annual Meeting of the Association for Computational Linguistics: Human Language Technologies*. Portland, Oregon, USA: Association for Computational Linguistics, June 2011, pp. 142–150. [Online]. Available: http://www.aclweb.org/anthology/P11-1015

[28] V. Sanh, L. Debut, J. Chaumond, and T. Wolf, "Distilbert, a distilled version of BERT: smaller, faster, cheaper and lighter," *CoRR*, vol. abs/1910.01108, 2019. [Online]. Available: http://arxiv.org/abs/1910.01108

[29] T. B. Brown, B. Mann, N. Ryder, M. Subbiah, J. Kaplan, P. Dhariwal, A. Neelakantan, P. Shyam, G. Sastry, A. Askell, S. Agarwal, A. Herbert-Voss, G. Krueger, T. Henighan, R. Child, A. Ramesh, D. M. Ziegler, J. Wu, C. Winter, C. Hesse, M. Chen, E. Sigler, M. Litwin, S. Gray, B. Chess, J. Clark, C. Berner, S. McCandlish, A. Radford, I. Sutskever, and D. Amodei, "Language models are few-shot learners," 2020.

[30] A. Radford, J. Wu, R. Child, D. Luan, D. Amodei, and I. Sutskever, "Language models are unsupervised multitask learners," 2019.

[31] S. Zhang, S. Roller, N. Goyal, M. Artetxe, M. Chen, S. Chen, C. Dewan, M. Diab, X. Li, X. V. Lin, T. Mihaylov, M. Ott, S. Shleifer, K. Shuster, D. Simig, P. S. Koura, A. Sridhar, T. Wang, and L. Zettlemoyer, "Opt: Open pre-trained transformer language models," 2022.

[32] H. Touvron, T. Lavril, G. Izacard, X. Martinet, M.-A. Lachaux, T. Lacroix, B. Rozière, N. Goyal, E. Hambro, F. Azhar, A. Rodriguez, A. Joulin, E. Grave, and G. Lample, "Llama: Open and efficient foundation language models," 2023.

[33] M. Hausknecht, P. Ammanabrolu, M.-A. Côté, and X. Yuan, "Interactive fiction games: A colossal adventure," *Proceedings of the AAAI Conference on Artificial Intelligence*, vol. 34, no. 05, pp. 7903–7910, Apr. 2020. [Online]. Available: https://ojs.aaai.org/index.php/AAAI/article/view/6297

[34] M. Ahn, A. Brohan, N. Brown, Y. Chebotar, O. Cortes, B. David, C. Finn, C. Fu, K. Gopalakrishnan, K. Hausman, A. Herzog, D. Ho, J. Hsu, J. Ibarz, B. Ichter, A. Irpan, E. Jang, R. J. Ruano, K. Jeffrey, S. Jesmonth, N. J. Joshi, R. Julian, D. Kalashnikov, Y. Kuang, K.-H. Lee, S. Levine, Y. Lu, L. Luu, C. Parada, P. Pastor, J. Quiambao, K. Rao, J. Rettinghouse, D. Reyes, P. Sermanet, N. Sievers, C. Tan, A. Toshev, V. Vanhoucke, F. Xia, T. Xiao, P. Xu, S. Xu, M. Yan, and A. Zeng, "Do as i can, not as i say: Grounding language in robotic affordances," 2022.

[35] B. R. Kiran, I. Sobh, V. Talpaert, P. Mannion, A. A. A. Sallab, S. Yogamani, and P. Pérez, "Deep reinforcement learning for autonomous driving: A survey," 2021.

[36] M. T. J. Spaan, *Partially Observable Markov Decision Processes*. Berlin, Heidelberg: Springer Berlin Heidelberg, 2012, pp. 387–414. [Online]. Available: https://doi.org/10.1007/978-3-642-27645-3_12

[37] V. Mnih, K. Kavukcuoglu, D. Silver, A. Graves, I. Antonoglou, D. Wierstra, and M. A. Riedmiller, "Playing atari with deep reinforcement learning," *CoRR*, vol. abs/1312.5602, 2013. [Online]. Available: http://arxiv.org/abs/1312.5602

[38] J. He, J. Chen, X. He, J. Gao, L. Li, L. Deng, and M. Ostendorf, "Deep reinforcement learning with a natural language action space," 2016.

[39] J. Chung, Ç. Gülçehre, K. Cho, and Y. Bengio, "Empirical evaluation of gated recurrent neural networks on sequence modeling," *CoRR*, vol. abs/1412.3555, 2014. [Online]. Available: http://arxiv.org/abs/1412.3555

[40] M. Côté, Á. Kádár, X. Yuan, B. Kybartas, T. Barnes, E. Fine, J. Moore, M. J. Hausknecht, L. E. Asri, M. Adada, W. Tay, and A. Trischler, "Textworld: A learning environment for text-based games," 2018. [Online]. Available: http://arxiv.org/abs/1806.11532

[41] J. He, J. Chen, X. He, J. Gao, L. Li, L. Deng, and M. Ostendorf, "Deep reinforcement learning with a natural language action space," in *Proceedings of the 54th Annual Meeting of the Association for Computational Linguistics (Volume 1: Long Papers)*. Berlin, Germany: Association for Computational Linguistics, Aug. 2016, pp. 1621–1630. [Online]. Available: https://aclanthology.org/P16-1153

[42] A. V. Sudhakar, P. Parthasarathi, J. Rajendran, and S. Chandar, "Language model-in-the-loop: Data optimal approach to learn-to-recommend actions in text games," 2023.

[43] A. Radford, J. Wu, R. Child, D. Luan, D. Amodei, and I. Sutskever, "Language models are unsupervised multitask learners," 2019. [Online]. Available: https://api.semanticscholar.org/CorpusID:160025533

[44] E. M. Bender and A. Koller, "Climbing towards NLU: On meaning, form, and understanding in the age of data," in *Proceedings of the 58th Annual Meeting of the Association for Computational Linguistics*. Online: Association for Computational Linguistics, Jul. 2020, pp. 5185–5198. [Online]. Available: https://aclanthology.org/2020.acl-main.463

[45] A. Rogers, O. Kovaleva, and A. Rumshisky, "A primer in bertology: What we know about how bert works," *Transactions of the Association for Computational Linguistics*, vol. 8, pp. 842–866, 2020.

[46] A. Webson and E. Pavlick, "Do prompt-based models really understand the meaning of their prompts?" *arXiv preprint arXiv:2109.01247*, 2021.

[47] Y. Bisk, A. Holtzman, J. Thomason, J. Andreas, Y. Bengio, J. Chai, M. Lapata, A. Lazaridou, J. May, A. Nisnevich, N. Pinto, and J. Turian,

"Experience grounds language," in *Proceedings of the 2020 Conference on Empirical Methods in Natural Language Processing (EMNLP)*. Online: Association for Computational Linguistics, Nov. 2020, pp. 8718–8735. [Online]. Available: https://aclanthology.org/2020.emnlp-main.703

[48] J. L. McClelland, F. Hill, M. Rudolph, J. Baldridge, and H. Schütze, "Placing language in an integrated understanding system: Next steps toward human-level performance in neural language models," *Proceedings of the National Academy of Sciences*, vol. 117, no. 42, pp. 25 966–25 974, 2020.

[49] B. M. Lake and G. L. Murphy, "Word meaning in minds and machines," *Psychological review*, 2021.

[50] S. Yao, R. Rao, M. Hausknecht, and K. Narasimhan, "Keep CALM and explore: Language models for action generation in text-based games," in *Proceedings of the 2020 Conference on Empirical Methods in Natural Language Processing (EMNLP)*. Online: Association for Computational Linguistics, Nov. 2020, pp. 8736–8754. [Online]. Available: https://aclanthology.org/2020.emnlp-main.704

[51] P. Ammanabrolu and M. Hausknecht, "Graph constrained reinforcement learning for natural language action spaces," in *International Conference on Learning Representations*, 2020. [Online]. Available: https://openreview.net/forum?id=B1x6w0EtwH

[52] I. Singh, G. Singh, and A. Modi, "Pre-trained language models as prior knowledge for playing text-based games," *CoRR*, vol. abs/2107.08408, 2021. [Online]. Available: https://arxiv.org/abs/2107.08408

[53] L. Chen, K. Lu, A. Rajeswaran, K. Lee, A. Grover, M. Laskin, P. Abbeel, A. Srinivas, and I. Mordatch, "Decision transformer: Reinforcement learning via sequence modeling," in *Advances in Neural Information Processing Systems*, A. Beygelzimer, Y. Dauphin, P. Liang, and J. W. Vaughan, Eds., 2021. [Online]. Available: https://openreview.net/forum?id=a7APmM4B9d

[54] M. Janner, Q. Li, and S. Levine, "Offline reinforcement learning as one big sequence modeling problem," in *Advances in Neural Information Processing Systems*, M. Ranzato, A. Beygelzimer, Y. Dauphin, P. Liang, and J. W. Vaughan, Eds., vol. 34. Curran Associates, Inc., 2021, pp. 1273–1286. [Online]. Available: https://proceedings.neurips.cc/paper/2021/file/099fe6b0b444c23836c4a5d07346082b-Paper.pdf

[55] Y. Xu, L. Chen, M. Fang, Y. Wang, and C. Zhang, "Deep reinforcement learning with transformers for text adventure games," in *2020 IEEE Conference on Games (CoG)*, 2020, pp. 65–72.

[56] E. Parisotto, H. F. Song, J. W. Rae, R. Pascanu, Ç. Gülçehre, S. M. Jayakumar, M. Jaderberg, R. L. Kaufman, A. Clark, S. Noury, M. M. Botvinick, N. Heess, and R. Hadsell, "Stabilizing transformers for reinforcement learning," *CoRR*, vol. abs/1910.06764, 2019. [Online]. Available: http://arxiv.org/abs/1910.06764

[57] L. Ouyang, J. Wu, X. Jiang, D. Almeida, C. Wainwright, P. Mishkin, C. Zhang, S. Agarwal, K. Slama, A. Gray, J. Schulman, J. Hilton, F. Kelton, L. Miller, M. Simens, A. Askell, P. Welinder, P. Christiano, J. Leike, and R. Lowe, "Training language models to follow instructions with human feedback," in *Advances in Neural Information Processing Systems*, A. H. Oh, A. Agarwal, D. Belgrave, and K. Cho, Eds., 2022. [Online]. Available: https://openreview.net/forum?id=TG8KACxEON

[58] M. Reid, Y. Yamada, and S. S. Gu, "Can wikipedia help offline reinforcement learning?" *CoRR*, vol. abs/2201.12122, 2022. [Online]. Available: https://arxiv.org/abs/2201.12122

[59] D. Tarasov, V. Kurenkov, and S. Kolesnikov, "Prompts and pre-trained language models for offline reinforcement learning," in *ICLR 2022 Workshop on Generalizable Policy Learning in Physical World*, 2022. [Online]. Available: https://openreview.net/forum?id=Spf4TE6NkWq

[60] M. Ahn, A. Brohan, N. Brown, Y. Chebotar, O. Cortes, B. David, C. Finn, C. Fu, K. Gopalakrishnan, K. Hausman, A. Herzog, D. Ho, J. Hsu, J. Ibarz, B. Ichter, A. Irpan, E. Jang, R. J. Ruano, K. Jeffrey, S. Jesmonth, N. J. Joshi, R. Julian, D. Kalashnikov, Y. Kuang, K.-H. Lee, S. Levine, Y. Lu, L. Luu, C. Parada, P. Pastor, J. Quiambao, K. Rao, J. Rettinghouse, D. Reyes, P. Sermanet, N. Sievers, C. Tan, A. Toshev, V. Vanhoucke, F. Xia, T. Xiao, P. Xu, S. Xu, M. Yan, and A. Zeng, "Do as i can, not as i say: Grounding language in robotic affordances," 2022. [Online]. Available: https://arxiv.org/abs/2204.01691

[61] J. Tuyls, S. Yao, S. M. Kakade, and K. R. Narasimhan, "Multi-stage episodic control for strategic exploration in text games," in *International Conference on Learning Representations*, 2022. [Online]. Available: https://openreview.net/forum?id=Ek7PSN7Y77z

[62] S. Li, X. Puig, C. Paxton, Y. Du, C. Wang, L. Fan, T. Chen, D. Huang, E. Akyürek, A. Anandkumar, J. Andreas, I. Mordatch, A. Torralba, and Y. Zhu, "Pre-trained language models for interactive decision-making," *arXiv*, 2022.

[63] J. Devlin, M. Chang, K. Lee, and K. Toutanova, "BERT: pre-training of deep bidirectional transformers for language understanding," in *Proceedings of the 2019 Conference of the North American Chapter of the Association for Computational Linguistics: Human Language Technologies, NAACL-HLT 2019, Minneapolis, MN, USA, June 2-7, 2019, Volume 1 (Long and Short Papers)*, J. Burstein, C. Doran, and T. Solorio, Eds. Association for Computational Linguistics, 2019, pp. 4171–4186. [Online]. Available: https://doi.org/10.18653/v1/n19-1423

[64] D. Khashabi, S. Min, T. Khot, A. Sabharwal, O. Tafjord, P. Clark, and H. Hajishirzi, "UNIFIEDQA: Crossing format boundaries with a single QA system," in *Findings of the Association for Computational Linguistics: EMNLP 2020*. Online: Association for Computational Linguistics, Nov. 2020, pp. 1896–1907. [Online]. Available: https://aclanthology.org/2020.findings-emnlp.171

[65] P. Rajpurkar, J. Zhang, K. Lopyrev, and P. Liang, "SQuAD: 100,000+ questions for machine comprehension of text," in *Proceedings of the 2016 Conference on Empirical Methods in Natural Language Processing*. Austin, Texas: Association for Computational Linguistics, Nov. 2016, pp. 2383–2392. [Online]. Available: https://aclanthology.org/D16-1264

[66] S. Y. Feng, V. Gangal, J. Wei, S. Chandar, S. Vosoughi, T. Mitamura, and E. Hovy, "A survey of data augmentation approaches for nlp," *arXiv preprint arXiv:2105.03075*, 2021.

[67] A. Radford, K. Narasimhan, T. Salimans, I. Sutskever *et al.*, "Improving language understanding by generative pre-training," 2018.

[68] Y. Wu, F. Li, and P. Liang, "Insights into pre-training via simpler synthetic tasks," *arXiv preprint arXiv:2206.10139*, 2022.

[69] P. Malviya and A. V. Sudhakar, "Feature diversity in self-supervised learning," 2022. [Online]. Available: https://arxiv.org/abs/2209.01275

[70] S. Welleck, I. Kulikov, S. Roller, E. Dinan, K. Cho, and J. Weston, "Neural text generation with unlikelihood training," *arXiv preprint arXiv:1908.04319*, 2019.

[71] Anonymous, "Multi-agent text-based hanabi challenge," in *Submitted to ICLR 2024 Workshop on Generative Models for Decision Making*, 2024, under review. [Online]. Available: https://openreview.net/forum?id=SGPUu7msST

[72] A. V. Sudhakar, H. Nekoei, M. Reymond, M. Liu, J. Rajendran, and S. Chandar, "A generalist hanabi agent," 2025. [Online]. Available: https://arxiv.org/abs/2503.14555

[73] T. Guo, X. Chen, Y. Wang, R. Chang, S. Pei, N. V. Chawla, O. Wiest, and X. Zhang, "Large language model based multi-agents: A survey of progress and challenges," 2024.

[74] S. Agashe, Y. Fan, and X. E. Wang, "Evaluating multi-agent coordination abilities in large language models," 2023.

[75] P. Zhou, A. Madaan, S. P. Potharaju, A. Gupta, K. R. McKee, A. Holtzman, J. Pujara, X. Ren, S. Mishra, A. Nematzadeh *et al.*, "How far are large language models from agents with theory-of-mind?" *arXiv preprint arXiv:2310.03051*, 2023.

[76] L. Ouyang, J. Wu, X. Jiang, D. Almeida, C. L. Wainwright, P. Mishkin, C. Zhang, S. Agarwal, K. Slama, A. Ray, J. Schulman, J. Hilton, F. Kelton, L. Miller, M. Simens, A. Askell, P. Welinder, P. Christiano, J. Leike, and R. Lowe, "Training language models to follow instructions with human feedback," 2022.

[77] A. Q. Jiang, A. Sablayrolles, A. Mensch, C. Bamford, D. S. Chaplot, D. de las Casas, F. Bressand, G. Lengyel, G. Lample, L. Saulnier, L. R. Lavaud, M.-A. Lachaux, P. Stock, T. L. Scao, T. Lavril, T. Wang, T. Lacroix, and W. E. Sayed, "Mistral 7b," 2023.

[78] N. Bard, J. N. Foerster, S. Chandar, N. Burch, M. Lanctot, H. F. Song, E. Parisotto, V. Dumoulin, S. Moitra, E. Hughes, I. Dunning, S. Mourad, H. Larochelle, M. G. Bellemare, and M. Bowling, "The hanabi challenge: A new frontier for AI research," *CoRR*, vol. abs/1902.00506, 2019. [Online]. Available: http://arxiv.org/abs/1902.00506

[79] D. Silver, A. Huang, C. J. Maddison, A. Guez, L. Sifre, G. van den Driessche, J. Schrittwieser, I. Antonoglou, V. Panneershelvam, M. Lanctot, S. Dieleman, D. Grewe, J. Nham, N. Kalchbrenner, I. Sutskever, T. P. Lillicrap, M. Leach, K. Kavukcuoglu, T. Graepel, and D. Hassabis, "Mastering the game of go with deep neural networks and tree search," *Nature*, vol. 529, pp. 484–489, 2016.

[80] J. Schulman, F. Wolski, P. Dhariwal, A. Radford, and O. Klimov, "Proximal policy optimization algorithms," *ArXiv*, vol. abs/1707.06347, 2017.

[81] M. Fang, C. Zhou, B. Shi, B. Gong, J. Xu, and T. Zhang, "Dher: Hindsight experience replay for dynamic goals," in *International Conference on Learning Representations*, 2018.

[82] M. Fang, Y. Li, and T. Cohn, "Learning how to active learn: A deep reinforcement learning approach," in *Conference on Empirical Methods in Natural Language Processing*, 2017.

[83] X. Yuan, M. Côté, A. Sordoni, R. Laroche, R. T. des Combes, M. J. Hausknecht, and A. Trischler, "Counting to explore and generalize in text-based games," *CoRR*, vol. abs/1806.11525, 2018. [Online]. Available: http://arxiv.org/abs/1806.11525

[84] X. Yin and J. May, "Learn how to cook a new recipe in a new house: Using map familiarization, curriculum learning, and bandit feedback to learn families of text-based adventure games," 2019. [Online]. Available: https://arxiv.org/abs/1908.04777

[85] K. Murugesan, M. Atzeni, P. Shukla, M. Sachan, P. Kapanipathi, and K. Talamadupula, "Enhancing text-based reinforcement learning agents with commonsense knowledge," *CoRR*, vol. abs/2005.00811, 2020. [Online]. Available: https://arxiv.org/abs/2005.00811

[86] R. Wang, P. A. Jansen, M.-A. Côté, and P. Ammanabrolu, "Scienceworld: Is your agent smarter than a 5th grader?" in *Conference on Empirical Methods in Natural Language Processing*, 2022.

[87] P. Osborne, H. Nõmm, and A. Freitas, "A survey of text games for reinforcement learning informed by natural language," *Transactions of the Association for Computational Linguistics*, vol. 10, pp. 873–887, 2022. [Online]. Available: https://aclanthology.org/2022.tacl-1.51

[88] E. M. Bender and A. Koller, "Climbing towards NLU: On meaning, form, and understanding in the age of data," in *Proceedings of the 58th Annual Meeting of the Association for Computational Linguistics*. Online: Association for Computational Linguistics, Jul. 2020, pp. 5185–5198. [Online]. Available: https://aclanthology.org/2020.acl-main.463

[89] J. He, J. Chen, X. He, J. Gao, L. Li, L. Deng, and M. Ostendorf, "Deep reinforcement learning with a natural language action space," *arXiv: Artificial Intelligence*, 2015.

[90] V. Jain, W. Fedus, H. Larochelle, D. Precup, and M. G. Bellemare, "Algorithmic improvements for deep reinforcement learning applied to interactive fiction," in *AAAI Conference on Artificial Intelligence*, 2019.

[91] Y. Xu, L. Chen, M. Fang, Y. Wang, and C. Zhang, "Deep reinforcement learning with transformers for text adventure games," *2020 IEEE Conference on Games (CoG)*, pp. 65–72, 2020.

[92] Y. Jang, S. Seo, J. Lee, and K.-E. Kim, "Monte-carlo planning and learning with language action value estimates," in *International Conference on Learning Representations*, 2021. [Online]. Available: https://openreview.net/forum?id=7_G8JySGecm

[93] S. Yao, R. Rao, M. Hausknecht, and K. Narasimhan, "Keep CALM and explore: Language models for action generation in text-based games," in *Proceedings of the 2020 Conference on Empirical Methods in Natural Language Processing (EMNLP)*, B. Webber, T. Cohn, Y. He, and Y. Liu, Eds. Online: Association for Computational Linguistics, Nov. 2020, pp. 8736–8754. [Online]. Available: https://aclanthology.org/2020.emnlp-main.704

[94] I. Singh, G. Singh, and A. Modi, "Pre-trained language models as prior knowledge for playing text-based games," *ArXiv*, vol. abs/2107.08408, 2021.

[95] D. Premack and G. Woodruff, "Does the chimpanzee have a theory of mind?" *Behavioral and Brain Sciences*, vol. 1, no. 4, p. 515–526, 1978.

[96] N. Rabinowitz, F. Perbet, F. Song, C. Zhang, S. M. A. Eslami, and M. Botvinick, "Machine theory of mind," in *Proceedings of the 35th International Conference on Machine Learning*, ser. Proceedings of Machine Learning Research, J. Dy and A. Krause, Eds., vol. 80. PMLR, 10–15 Jul 2018, pp. 4218–4227. [Online]. Available: https://proceedings.mlr.press/v80/rabinowitz18a.html

[97] T. Ullman, "Large language models fail on trivial alterations to theory-of-mind tasks," *arXiv preprint arXiv:2302.08399*, 2023.

[98] M. F. A. R. D. T. (FAIR)†, A. Bakhtin, N. Brown, E. Dinan, G. Farina, C. Flaherty, D. Fried, A. Goff, J. Gray, H. Hu, A. P. Jacob, M. Komeili, K. Konath, M. Kwon, A. Lerer, M. Lewis, A. H. Miller, S. Mitts, A. Renduchintala, S. Roller, D. Rowe, W. Shi, J. Spisak, A. Wei, D. Wu, H. Zhang, and M. Zijlstra, "Human-level play in the game of <i>diplomacy</i> by combining language models with strategic

reasoning," *Science*, vol. 378, no. 6624, pp. 1067–1074, 2022. [Online]. Available: https://www.science.org/doi/abs/10.1126/science.ade9097

[99] J. Foerster, F. Song, E. Hughes, N. Burch, I. Dunning, S. Whiteson, M. Botvinick, and M. Bowling, "Bayesian action decoder for deep multi-agent reinforcement learning," in *International Conference on Machine Learning.* PMLR, 2019, pp. 1942–1951.

[100] H. Hu and J. N. Foerster, "Simplified action decoder for deep multi-agent reinforcement learning," *arXiv preprint arXiv:1912.02288*, 2019.

[101] H. Nekoei, A. Badrinaaraayanan, A. Courville, and S. Chandar, "Continuous coordination as a realistic scenario for lifelong learning," in *International Conference on Machine Learning.* PMLR, 2021, pp. 8016–8024.

[102] A. Lupu, B. Cui, H. Hu, and J. Foerster, "Trajectory diversity for zero-shot coordination," in *Proceedings of the 38th International Conference on Machine Learning*, ser. Proceedings of Machine Learning Research, M. Meila and T. Zhang, Eds., vol. 139. PMLR, 18–24 Jul 2021, pp. 7204–7213. [Online]. Available: https://proceedings.mlr.press/v139/lupu21a.html

[103] H. Hu, A. Lerer, B. Cui, L. Pineda, N. Brown, and J. Foerster, "Off-belief learning," in *Proceedings of the 38th International Conference on Machine Learning*, ser. Proceedings of Machine Learning Research, M. Meila and T. Zhang, Eds., vol. 139. PMLR, 18–24 Jul 2021, pp. 4369–4379. [Online]. Available: https://proceedings.mlr.press/v139/hu21c.html

[104] B. Cui, H. Hu, L. Pineda, and J. Foerster, "K-level reasoning for zero-shot coordination in hanabi," in *Advances in Neural Information Processing Systems*, M. Ranzato, A. Beygelzimer, Y. Dauphin, P. Liang, and J. W. Vaughan, Eds., vol. 34. Curran Associates, Inc., 2021, pp. 8215–8228. [Online]. Available: https://proceedings.neurips.cc/paper/2021/file/4547dff5fd7604f18c8ee32cf3da41d7-Paper.pdf

[105] K. Lucas and R. E. Allen, "Any-play: An intrinsic augmentation for zero-shot coordination," 2022. [Online]. Available: https://arxiv.org/abs/2201.12436

[106] H. Nekoei, X. Zhao, J. Rajendran, M. Liu, and S. Chandar, "Towards few-shot coordination: Revisiting ad-hoc teamplay challenge in the game of hanabi," in *Conference on Lifelong Learning Agents.* PMLR, 2023, pp. 861–877.

[107] H. Hu and D. Sadigh, "Language instructed reinforcement learning for human-ai coordination," 2023.

[108] V. Sanh, L. Debut, J. Chaumond, and T. Wolf, "Distilbert, a distilled version of BERT: smaller, faster, cheaper and lighter," *CoRR*, vol. abs/1910.01108, 2019. [Online]. Available: http://arxiv.org/abs/1910.01108

[109] R. S. Sutton and A. G. Barto, *Reinforcement Learning: An Introduction*, 2nd ed. Cambridge, MA, USA: The MIT Press, 2018. [Online]. Available: http://incompleteideas.net/book/the-book-2nd.html

[110] E. J. Hu, Y. Shen, P. Wallis, Z. Allen-Zhu, Y. Li, S. Wang, L. Wang, and W. Chen, "Lora: Low-rank adaptation of large language models," 2021.

[111] M. Hessel, J. Modayil, H. van Hasselt, T. Schaul, G. Ostrovski, W. Dabney, D. Horgan, B. Piot, M. Azar, and D. Silver, "Rainbow: Combining improvements in deep reinforcement learning," 2017.

[112] T. Wolf, L. Debut, V. Sanh, J. Chaumond, C. Delangue, A. Moi, P. Cistac, T. Rault, R. Louf, M. Funtowicz, J. Davison, S. Shleifer, P. von Platen, C. Ma, Y. Jernite, J. Plu, C. Xu, T. Le Scao, S. Gugger, M. Drame, Q. Lhoest, and A. Rush, "Transformers: State-of-the-art natural language processing," in *Proceedings of the 2020 Conference on Empirical Methods in Natural Language Processing: System Demonstrations*. Online: Association for Computational Linguistics, Oct. 2020, pp. 38–45. [Online]. Available: https://aclanthology.org/2020.emnlp-demos.6

[113] A. Joulin, E. Grave, P. Bojanowski, and T. Mikolov, "Bag of tricks for efficient text classification," in *Proceedings of the 15th Conference of the European Chapter of the Association for Computational Linguistics: Volume 2, Short Papers*. Valencia, Spain: Association for Computational Linguistics, Apr. 2017, pp. 427–431. [Online]. Available: https://aclanthology.org/E17-2068

[114] L. Biewald, "Experiment tracking with weights and biases," 2020, software available from wandb.com. [Online]. Available: https://www.wandb.com/

[115] H. Hu, A. Lerer, B. Cui, L. Pineda, D. J. Wu, N. Brown, and J. N. Foerster, "Off-belief learning," *CoRR*, vol. abs/2103.04000, 2021. [Online]. Available: https://arxiv.org/abs/2103.04000

[116] ChandarLab, "RLHive," https://github.com/chandar-lab/RLHive?tab=readme-ov-file#installing, 2021.

# APPENDIX A   LANGUAGE MODEL IN THE LOOP: TOWARDS ADAPTATION

## A.1   Language Model Setup

We use a GPT-2 (Base) [43] model with 12-layers, 768-hidden units, and 12- attention heads with 117M parameters pre-trained on the WebText corpus. This model's implementation and pretrained weights are obtained from [112, Huggingface].

We train for 3 epochs on the ClubFloyd dataset following [50] to minimize the cross-entropy loss, as shown in Table A.1. We use AdamW to optimize model's weights to minimize the loss, with the learning rate as $2 \times 10^{-6}$ and Adam epsilon as $1 \times 10^{-9}$. We use a linear schedule with a warmup of 0.1 for the learning rate. Finally, we clip gradients with a maximum gradient norm of 1. Following [50]'s finetuning process, we exclude using Jericho-related transcripts by setting the flag as 1. We used random seeds to select the dataset to avoid bias in selecting data for the LM training.

| Model | Metric | Final Score(3 epoch) |
|-------|--------|----------------------|
| 100%  | Train Loss | 1.49 |
|       | Val Loss | 2.65 |
|       | Train Acc | 0.30 |
|       | Val Acc | 0.14 |
| 10%   | Train Loss | 1.42 |
|       | Val Loss | 3.04 |
|       | Train Acc | 0.30 |
|       | Val Acc | 0.09 |

Table A.1 Pre-trained GPT-2 Language Model training details on different data percentage variants trained.

## A.2   Reinforcement Learning Agent Setup:

We train on 10 interactive fiction games from the Jericho benchmark [33]. The states are observations concatenated with items in possession of the player and their current location description provided by the game engine using commands inventory and look. A single game episode runs for 100 environment steps at max or gets terminated before the game is over or won. We use the `look` and `inventory` commands to add location and inventory descriptions to observations, following [33].

We train DRRN asynchronously on 8 parallel instances of the game environment for $100,000$

steps for each game. At each step, the Q-value is estimated using the DRRN agent, and the action is selected based on the soft-exploration policy. Action's admissibility is predicted based on the textual response of the game. Then, inadmissible are filtered out using a FastText model [113]. The agent is optimized using adam optimizer with a learning rate of $10^{-5}$. We sample transitions of batch size 64 from priority buffer with a priority fraction of 0.5. The discount factor in determining the future reward's importance is 0.9. The size of the embedding dimension is 128, and the hidden dimension is 128. Finally, the gradient is clipped with a maximum gradient norm of 5. We train 5 separate runs for each game and report the average score. We use the average of the last 100 episode scores to calculate the final score.

## A.3   Acceleration plots

Comparison of learning dynamics of the different LM-in-the-Loop techniques with the baseline CALM agent across the selected 10 games in Jericho.

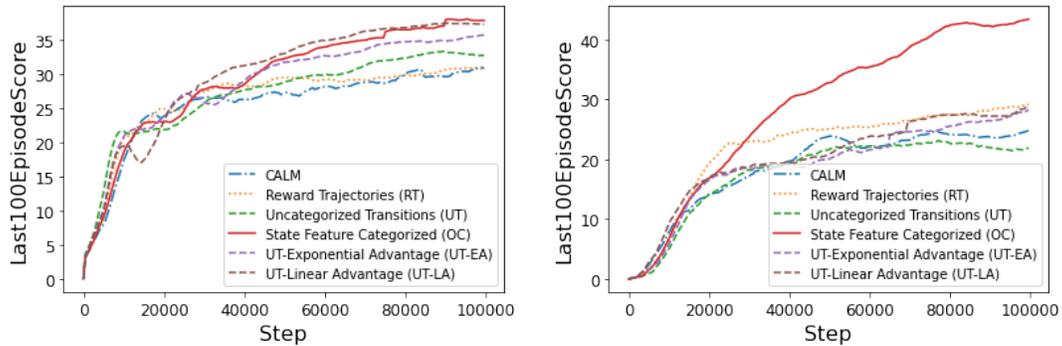

Figure A.1 a) zork-1 b) inhumane

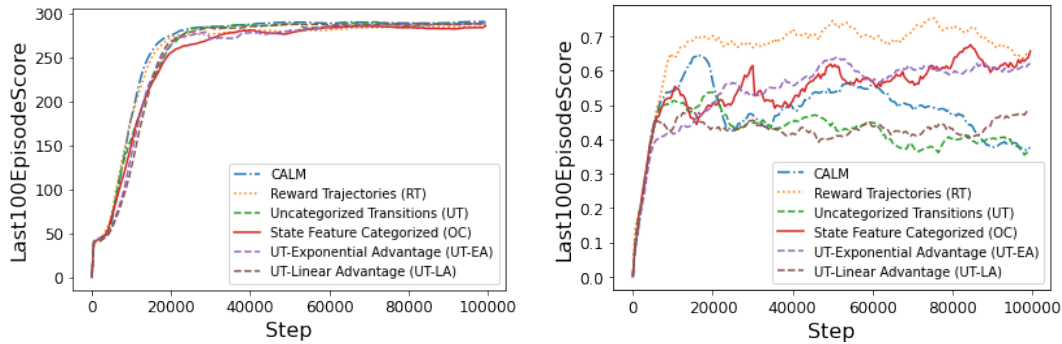

Figure A.2 c) detective d) zork3

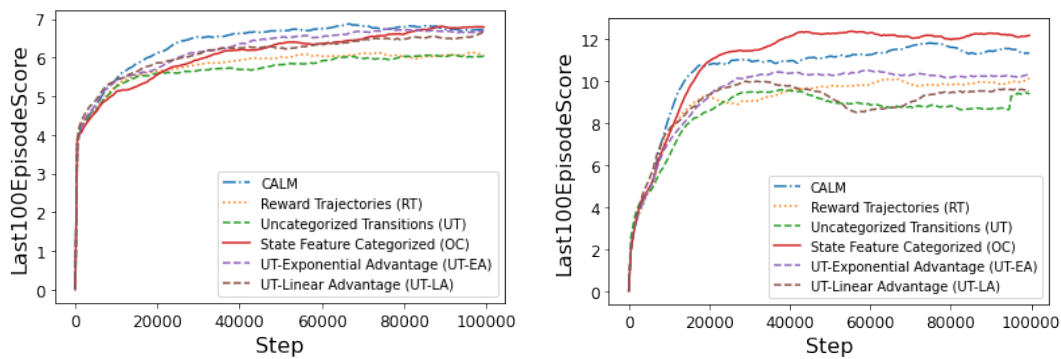

Figure A.3 e) omniquest f) library

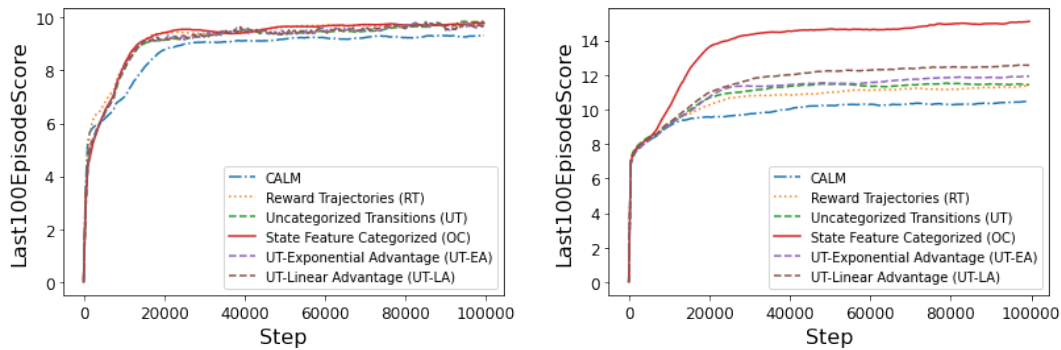

Figure A.4 g) balances h) ludicorp

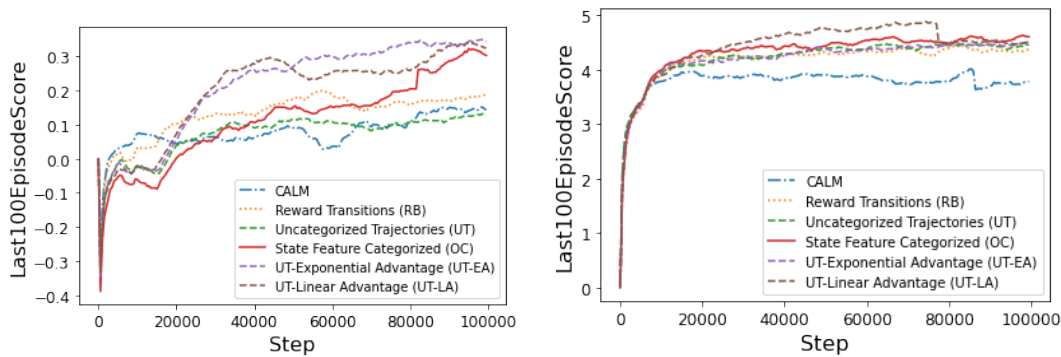

Figure A.5 i) dragon j) ztuu

## APPENDIX B    LM IN A MULTI-AGENT SETUP

### B.1    How Good LLMs are in playing Hanabi?

To adapt the LLaMA to the gameplay, we use Low-Rank Adaptation, or LoRA [110], which learns a low-rank decomposition matrices into each layer of the transformer architecture and freezes the pre-trained model weights. Thereby, significantly reducing the trainable parameters. We conducted fine-tuning experiments with LLaMA-7B weights with classifier using varying data sizes $[200, 500, 1000]$ and LoRA ranks $[32, 64, 128]$ for 10 epoch. Despite these parameter variations, the gameplay scores remained suboptimal level of around one as shown in B.1. This highlights the challenges in achieving effective gameplay performance for current large langue model on playing hanabi.

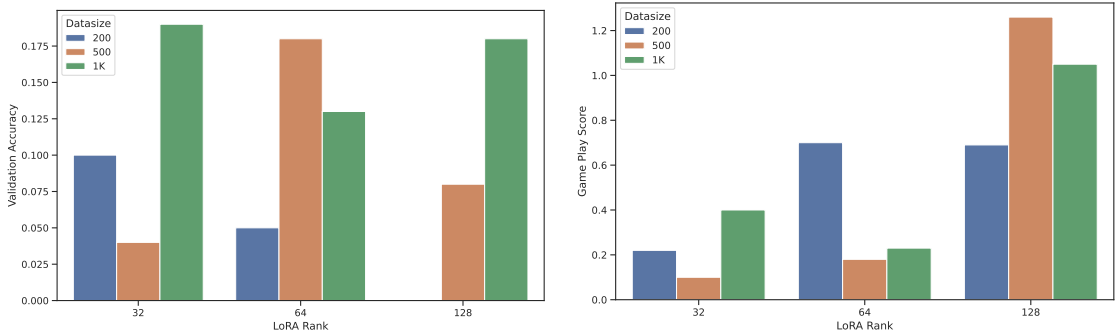

Figure B.1 *Evaluation of Low-Rank Adaptation (LoRA) in LLaMA-7B finetuning, showcasing the impact on a) Validation Accuracy and b) Game Play Score. The experiments involve varying data sizes [200, 500, 1000] and LoRA ranks [32, 64, 128].*

### B.2    Ablation studies

#### B.2.1    The role of scaling the dataset and different model variants

The dataset size emerges as a pivotal factor influencing gameplay scores. As the amount of training data increases there is a gradual increase in validation and the gameplay score. When the training percentage is equal to or less than 10% the games scores were poor ranging around 1 out of 25. In contrast, the gameplay score sharply increases when using 25% of the data as shown in B.2b. Nevertheless, the performance plateaus at a game play score of approximately 9 for both 75% and 100% , indicative of reaching a saturation point, affirming the sufficiency of the dataset size for effective model training.

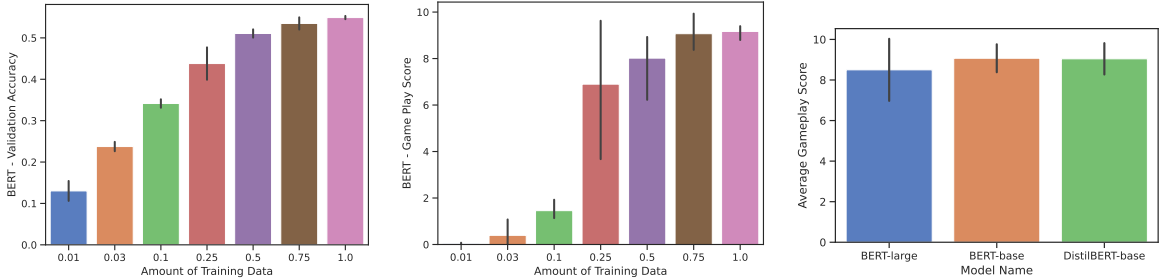

Figure B.2 *Analysis of the impact of training data amount on BERT, examining a) BERT Validation Accuracy, b) BERT Game Play Score across different percentages of training data, and c) BERT model variants with varying parameter sizes.*

In our experimentation, we varied the model parameter sizes—ranging from DistilBERT with $66M$ parameters to BERT-base-uncased with $110M$ parameters and BERT-large-uncased with $340M$ parameters. We observed that DistilBERT achieves a competitive gameplay score of approximately 8.7 after 600 game runs B.2c. On top of the performance considering the fast inference and low memory usage, DistilBERT was chosen as a candidate for integration with reinforcement learning through distillation.

### B.2.2   The role of discard information

We examined the impact of incorporating the discard pile into the observation. Surprisingly, we discovered that utilizing the discard pile did not contribute to any improvement in game scores as show in the Figure B.3. Rather, it resulted in a doubling of the sequence length of the language model. Given the need for fast inference in the reinforcement learning pipeline, we opted to exclude discard pile information from the observation during both language model training and inference. Nonetheless, there is a potential for heuristic-based approaches, to explore the idea of creating derived information from from the discard pile, potentially leading to a more concise sequence length and better game score.

## B.3   Training details

### B.3.1   Language Model setup

The model's finetuning process begins with a set of training instances, denoted as $(S, A)$ drawn from the dataset $\mathbb{D}$ where $S \in \{s_0, s_1, .., s_n\}$ and $A \in \{a_0, a_1, .., a_n\}$. Within this set, $s$ and $a$ represent a state and its corresponding noisy labelled action, respectively, and $n$ represents the number of examples in the dataset. The training objective of BERT, DistilBERT,

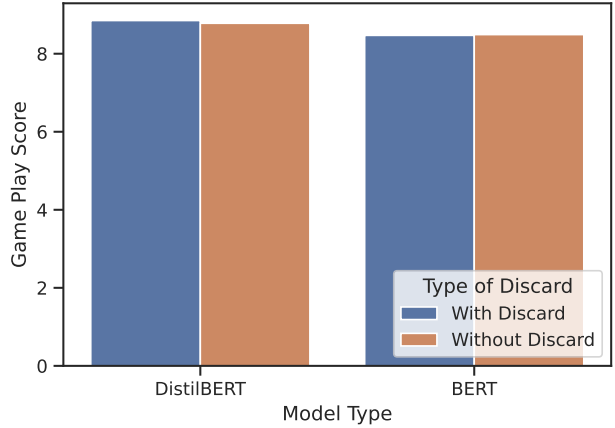

Figure B.3 *Evaluation of the discard pile's role in the game is assessed by comparing game scores with the presence and absence of the discard pile in the observation during training.*

GPT2-Classifier is,

$$L_{CCE} = -\frac{1}{N} \sum_{i=1}^{N} \sum_{j=1}^{C} a_{ij} \log(\hat{a}_{ij}) \qquad (B.1)$$

Where $N$ is the batch size. $C$ is the number of classes. $a_{ij}$ is the true probability of class j for the i-th example in the batch and $\hat{a}_{ij}$ is the predicted probability of class j for the i-th example in the batch.

The training objective of GPT-2 Generative is to minimize the cross-entropy loss, denoted as $\mathcal{L}$, and do the finetuning of the model. The cross-entropy loss is mathematically defined as follows:

$$\mathcal{L}_{LLM} = -\mathbb{E}_{(S,A)\sim D} \log p(A|S) \qquad (B.2)$$

Where $p(S|A)$ represents the conditional probability of predicting an action $A$, given the state $S$. The goal is to optimize these parameters, by minimizing the cross-entropy loss. We finetune the model to generate responses that better align with Hanabi game. The learning graph of validation accuracy with the game play score for each epoch is logged to understand the trend in the Figure B.4(a,b). Mostly the Validation score and game score is getting saturated at around 4th epoch.

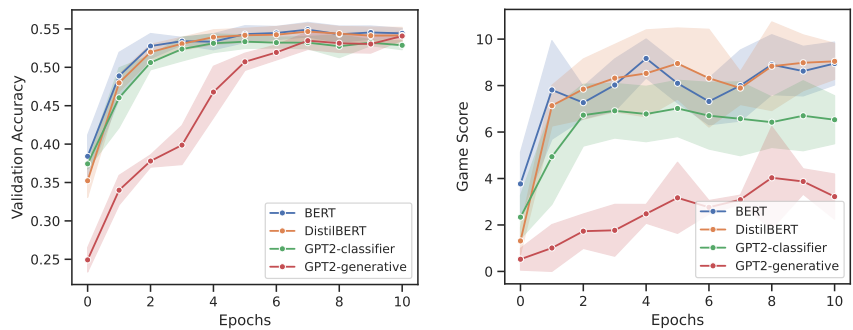

Figure B.4 Learning graph for (a) Validation accuracy plotted against(b) Game play score, for each epoch for different language model providing insights into the observed trends during the training process.

### B.3.2  Software details

The code was implemented using PyTorch, and pre-trained language models were loaded using Huggingface. To gain insights for this paper, we employed Weights & Biases [114] for experiment tracking and visualizations. Lastly, plots are created using the seaborn package. For RL algorithms, we used OBL agent [115] to collect the expert trajectory and RL Hive [116] to train the algorithm.